\documentclass[runningheads]{llncs}

\usepackage{eccv}

\usepackage{eccvabbrv}

\usepackage{graphicx}
\usepackage{booktabs}
\usepackage{array}
\usepackage[table]{xcolor} 
\usepackage{multirow}
\usepackage{placeins}
\usepackage{adjustbox}
\usepackage{fix-cm}
\usepackage{siunitx}
\usepackage{tabularx}   
\usepackage{makecell} 

\usepackage{microtype}
\usepackage{tcolorbox}
\tcbuselibrary{breakable}
\usepackage{comment}
\usepackage{algorithm}
\usepackage{algpseudocode}
\usepackage{pifont}

\usepackage[accsupp]{axessibility}  

\usepackage{hyperref}

\usepackage{orcidlink}
\definecolor{darkgreen}{RGB}{0, 150, 0}

\begin{document}

\title{Ego-Human Motion Prediction \\with 3D-Aware LLM}

\titlerunning{Ego3DLM}

\newcommand\CoAuthorMark{\footnotemark[\arabic{footnote}]}

\author{Yujin Bae\orcidlink{0009-0002-9225-8588}\thanks{Equal contribution.} \and
Jaewoo Jeong\orcidlink{0000-0001-8789-792X}\protect\CoAuthorMark \and
Hyeonseong Kim\orcidlink{0009-0003-9792-4647}\protect\CoAuthorMark \and
Kuk-Jin Yoon\orcidlink{0000-0002-1634-2756}}

\authorrunning{Bae et al.}

\institute{Visual Intelligence Lab., KAIST, Korea\\
\email{\{yujinbae,jeong207,brian617,kjyoon\}@kaist.ac.kr}}

\maketitle

\begin{abstract}
Anticipating human motion from an egocentric perspective is fundamental for proactive assistance in AR/VR, human-robot collaboration, and embodied AI. While recent works incorporate language as a semantic prior to reduce the ill-posed nature of egocentric forecasting, they largely neglect the 3D spatial and semantic context that governs how motion unfolds, and treat pose and language prediction as separate inference streams.
We introduce Ego3DLM, built on two core principles: accurate motion forecasting requires explicit spatial and semantic understanding of the 3D environment, and pose and language must be predicted holistically in a single pass, since motion is inherently tied to the semantic interpretation of actions being performed. 
Given three-point tracking, 3D scene features, and egocentric video, Ego3DLM simultaneously decodes past pose, future pose, past narration, and future narration in a single autoregressive pass, grounding predicted poses and descriptions in one another to enforce cross-modal and temporal consistency. We adopt a three-stage training scheme: (1) spatial-semantic scene awareness pretraining; (2) holistic instruction tuning over all four outputs in a single pass; and (3) GRPO-based reinforcement finetuning with intra- and inter-modal rewards that directly optimize pose-language fidelity. Experiments on the Nymeria benchmark demonstrate that Ego3DLM achieves state-of-the-art performance across future motion prediction, past motion tracking, and motion description, showing that 3D scene grounding and holistic cross-modal prediction yield physically plausible and semantically coherent motion forecasts. The project page is available at \url{https://jaewoo97.github.io/Ego3DLM/}.
\end{abstract}

\section{Introduction}
\label{sec:intro}

\begin{figure}
    \centering
    \includegraphics[width=0.8\linewidth]{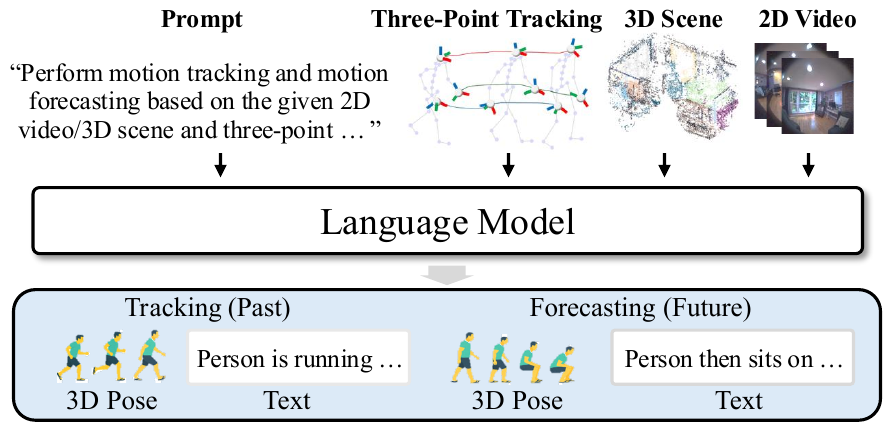}
    \caption{
    Our Ego3DLM incorporates the semantic context of the surrounding 3D environment along with 2D egocentric video and three-point tracking data to generate past and future poses and their corresponding language motion descriptions. 
    }
    \label{fig:teaser}
\end{figure}

Predicting human motion from an egocentric perspective is fundamental for proactive, real-time assistance in AR/VR, human–robot collaboration, and embodied AI~\cite{mao2022contact, guo2023back, mao2021generating, yue2024human, wang2024harmonizing, peng2023trajectory, xu2023eqmotion, xu2023joint}. Unlike third-person footage, egocentric observations reflect the wearer's true field of view and thus encode intent, affordances, and near-field interactions that matter for planning~\cite{frischen2007gaze, friesen1998eyes, land2001ways}. 
Yet forecasting from this viewpoint is inherently ill-posed: the camera captures little of the body, only sparse wearable cues are available, and the same partial observation is consistent with many plausible future motions. 
Resolving this ambiguity requires grounding predictions in semantic context: understanding not merely the kinematic state of the body, but the underlying intent and action semantics that govern how motion will unfold. 

Towards this end, recent works have sought to reduce forecast ambiguity by incorporating language as a semantic prior on human intention, demonstrating that treating motion as a tokenized sequence enables unified reasoning across vision, language, and pose~\cite{jiang2023motiongpt, hong2025egolm}.
Motion-language models such as MotionGPT~\cite{jiang2023motiongpt} and EgoLM~\cite{hong2025egolm} cast 3D human motion as ``motion tokens'' for joint modeling with text, while pose-language systems such as ChatPose~\cite{feng2024chatpose} and PoseScript~\cite{delmas2024posescript} demonstrate that large language models can understand and generate fine-grained 3D human poses from textual descriptions. However, most existing methods emphasize reconstruction, captioning, or editing of observed motion rather than prediction, and often operate on third-person videos or canonicalized motion capture rather than egocentric inputs. 

Forecasting works on egocentric motion, such as EgoCast~\cite{escobar2025egocast} and UniEgoMotion~\cite{patel2025uniegomotion}, focus on pose prediction from egocentric videos and sparse proprioceptive signals. 
However, these methods typically omit explicit 3D scene representations whose spatial and semantic context is crucial for resolving the ambiguity of future motion in cluttered, contact-rich environments. 
While FIction~\cite{ashutosh2025fiction} and $\text{HMD}^2$~\cite{guzov2025hmd} employ an explicit 3D scene representation for pose prediction and motion generation, they inject these features directly into the model, which does not guarantee that the model acquires genuine 3D reasoning capabilities.
Furthermore, prior works treat pose and language as separate inference streams, forgoing the cross-modal consistency that arises from holistic predictions and missing the opportunity to leverage semantic reasoning as a direct inductive bias during forecasting.

We address these limitations with Ego3DLM, built on two core principles. First, accurate human motion prediction requires explicit spatial and semantic understanding of the 3D environment, which defines the physical constraints, free space, and object-level affordances that govern possible movements. Second, pose and language must be predicted holistically in a single pass rather than independently, since human motion is inherently tied to the semantic interpretation of the actions being performed. To this end, our model takes three-point tracking (\ie head and hand poses), 3D scene features, and egocentric video as input, and simultaneously decodes past pose, future pose, past text, and future text in a single forward pass (Fig.~\ref{fig:teaser}), grounding predicted poses and descriptions in one another to improve cross-modal and temporal consistency.

These principles directly motivate our three-stage training scheme. In Stage I, we ground the language model (LM) in the 3D environment by pretraining on automatically generated spatial and semantic question-answer (QA) pairs from 3D scenes, instilling obstacle awareness and object-level semantics before introducing motion reasoning. In Stage II, the scene-aware LM is instruction-tuned to holistically generate all four outputs in a single autoregressive pass, with an explicit spatial scene reasoning step prepended so that spatial understanding propagates through pose prediction and into language narration. In Stage III, Group Relative Policy Optimization (GRPO) with intra- and inter-modal rewards directly optimizes for pose and language accuracy and their fidelity, which likelihood-based training alone cannot enforce. This stage fully leverages cross-modal coherence between pose and language to encourage the model toward predictions that are both physically plausible and semantically consistent.

We validate Ego3DLM on the Nymeria~\cite{ma2024nymeria} egocentric motion benchmark, which provides synchronized egocentric videos, language descriptions, and reconstructed 3D scenes. 
Our model achieves state-of-the-art performance across all metrics for future motion prediction and past motion tracking, along with their corresponding language descriptions, demonstrating comprehensive cross-modal understanding of human motion and intent.
Our contributions are three-fold:

\begin{itemize}
\item \textbf{3D scene-aware egocentric motion prediction:} a unified framework that jointly forecasts language and 3D pose for both past and future from three-point cues, egocentric video, and explicit 3D scene features.
\item \textbf{Simultaneous four-output decoding:} a single-pass generation of past pose, past text, future pose, and future text that enforces temporal and cross-modal coherence, improving fidelity between predicted motion and language.
\item \textbf{Multi-stage training:} a three-stage training scheme that progressively builds from 3D spatial-semantic scene understanding to holistic egocentric motion tracking and forecasting with joint pose-language supervision.
\end{itemize}

\section{Related Works}

\subsection{Multi-Modal Language Models}
Recent large language models (LLMs) have expanded beyond text and RGB images to operate on diverse modalities such as event streams~\cite{wu2023eventclip,liu2025eventgpt, zhou2024exact, zhou2024eventbind, qin2025event}, point clouds\cite{hong20233d, fu2024scene, chen2024ll3da, cao2023coda, delitzas2023multi, xu2024pointllm, zheng2025video, huang2024chat, kang2025robin3d}, and 3D human pose~\cite{jiang2023motiongpt, tevet2022motionclip, chen2025motionllm, zhang2024large, feng2024chatpose, zhou2024avatargpt, fan2025go}. EventGPT~\cite{liu2025eventgpt} extends LLMs to event cameras via spatio-temporal aggregation, while 3D-oriented models align point cloud encoders with language models~\cite{chen2024ll3da, hong20233d, fu2024scene}. Other approaches~\cite{zhou2024avatargpt, jiang2023motiongpt, hong2025egolm} tokenize SMPL~\cite{loper2023smpl} parameters as pose tokens~\cite{van2017neural, huang2024pq}, and EgoLM~\cite{hong2025egolm} models the joint distribution of egocentric motion and language for wearable settings. Moreover, discrete motion tokens have been treated as a foreign “language”, enabling unified motion-language reasoning~\cite{jiang2023motiongpt}. While these works show that heterogeneous inputs (events, points, pose) can be projected into a shared token space, most focus on understanding or reconstruction, with limited attention to forecasting multi-modal futures from explicit 3D scene geometry. Building upon them, we incorporate 3D scene-conditioned forecasting to predict human motion guided by semantic objectives.

\subsection{Human Motion Understanding \& Forecasting} 
Text-motion models~\cite{tevet2022human, athanasiou2022teach, guo2022tm2t, zhang2024motiondiffuse, zhang2023generating, zhang2023remodiffuse, pinyoanuntapong2024bamm, xie2023omnicontrol, pinyoanuntapong2024mmm, guo2022generating } have explored bidirectional mappings between natural language and 3D human motion. TEACH~\cite{athanasiou2022teach} composes long motions from paragraphs by stitching atomic actions, while diffusion-based models such as MDM~\cite{tevet2022human} and MotionDiffuse~\cite{zhang2024motiondiffuse} generate realistic and diverse motions from text by viewing motion synthesis as a denoising process in pose space. MotionGPT~\cite{jiang2023motiongpt} unifies multiple motion-language tasks (captioning, text-driven generation, prediction) under a single LLM-style architecture by discretizing motion into tokens. These approaches show that language supervision can ground high-level semantics of human motion, but they typically operate in third-person views.
EgoLM~\cite{hong2025egolm} extends motion-language modeling to the egocentric setting by jointly leveraging egocentric video and three-point tracking data. Nevertheless, these works do not explicitly account for the surrounding 3D scene, limiting their ability to anticipate scene-dependent contacts and physical constraints.

Human motion forecasting targets future 3D body pose~\cite{jeong2024multi, bouazizi2022motionmixer, chen2023humanmac, choudhury2023tempo, mao2022contact, rahman2023best, saadatnejad2023generic, wang2021simple, zhou2024smartrefine, yue2024human, mao2019learning}, hand pose, or trajectory~\cite{aydemir2023adapt, choi2023r, jiang2023motiondiffuser, mao2023leapfrog, park2024t4p, jeong2025multi, qiu2025adapting, bahari2025certified}. In the egocentric setting, EgoCast~\cite{escobar2025egocast} and UniEgoMotion~\cite{patel2025uniegomotion} forecast 3D body pose, while hand-centric models such as HOI-Forecast~\cite{liu2022joint} and EgoH4~\cite{hatano2025invisible} predict 3D hand motion and contact hotspots. Yet these methods typically predict either pose or hand trajectories alone and lack rich language descriptions. A separate line of work, FIction~\cite{ashutosh2025fiction} and HMD$^2$~\cite{guzov2025hmd}, grounds motion prediction and tracking in 3D scenes, but injects scene geometry directly and encodes it implicitly, without explicit scene reasoning. In contrast, our model jointly predicts 3D pose and textual descriptions of both past and future from egocentric video and sparse three-point tracking, while reasoning explicitly over 3D scene geometry. This encourages diverse yet physically and semantically coherent motion-language futures that respect scene layout, obstacles, and affordances.

\section{Method}
\begin{figure*}[t]
    \centering
    \includegraphics[width=1.0\linewidth]{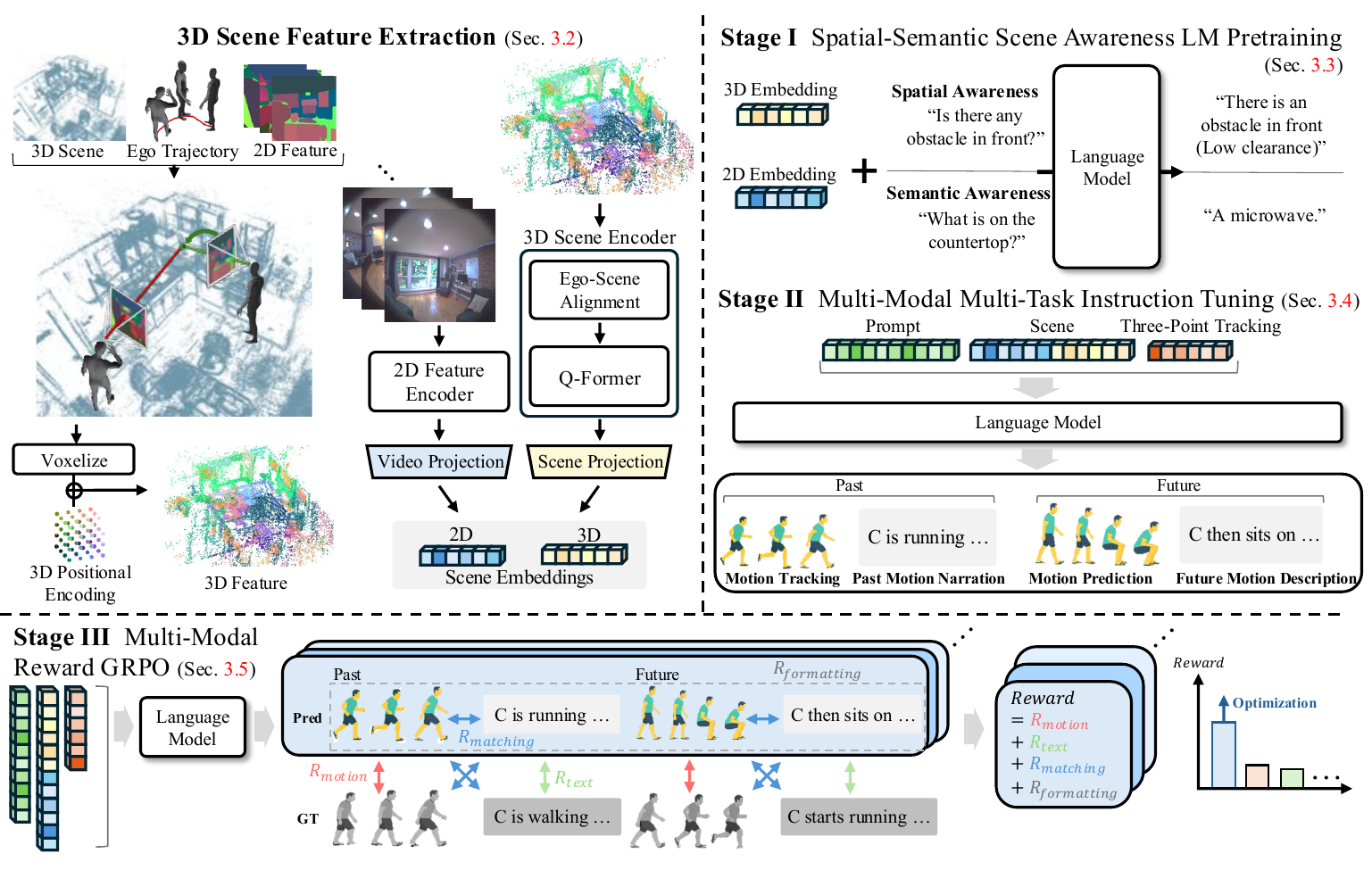} 
    \caption{
        \textbf{Overview of the Ego3DLM framework.}
        Our model consists of three stages.
        \textbf{3D Scene Feature Extraction} (Sec.~\ref{sec:scene}): 2D semantic features are extracted from egocentric video frames and lifted onto the 3D point cloud. The feature-enhanced point cloud is fed into a Q-Former to produce compact scene query embeddings (top left).
        \textbf{LM Training:} \textbf{(I) Pre-training} (Sec.~\ref{sec:pretrain}) aligns scene, motion, and language via two complementary objectives;  \textbf{(II) Instruction Tuning} (Sec.~\ref{sec:tuning}) trains simultaneous single-pass generation of all four outputs, comprising past and future poses and descriptions (top right); \textbf{(III) GRPO} (Sec.~\ref{sec:grpo}) refines the model with intra- and inter-modal rewards for pose-language fidelity (bottom).
    }
    \label{fig:overview}
\end{figure*}

We posit that predicting future human motion requires spatial and semantic understanding of the surrounding 3D environment, which defines the physical constraints and affordances of possible movements.
Furthermore, human motion is inherently tied to the semantic interpretation of actions being performed, necessitating joint reasoning over motion and language rather than treating them independently. 
Grounded in this principle, we introduce Ego3DLM trained on three stages, as shown in Fig.~\ref{fig:overview}. Spatial-Semantic Scene Awareness LM Pretraining (Stage I) encodes 3D scene context and injects it into the LM (\cref{sec:scene}) and trains spatial and semantic scene awareness (\cref{sec:pretrain}). Multi-Modal Multi-Task Instruction Tuning (Stage II) trains holistic motion tracking and prediction alongside past motion narration and future motion description (\cref{sec:tuning}). Finally, Multi-Modal Reward GRPO (Stage III) boosts the accuracy and cross-modal coherence of generated pose and text (\cref{sec:grpo}).

\subsection{Preliminaries}
\label{sec:pre}
\noindent\textbf{Problem Formulation.}
Given three-point motion tracking features from 6D poses $\mathbf{P}$, egocentric video embeddings $\mathbf{V}$, and 3D scene embeddings $\mathbf{S}$, our Ego3DLM jointly predicts four outputs in a single autoregressive pass: past pose tokens $x^{\text{past}}$, future pose tokens $x^{\text{fut}}$, past motion narration $y^{\text{past}}$, and future motion description $y^{\text{fut}}$. 
Each pose sequence $x \in \{x^{\text{past}}, x^{\text{fut}}\}$ is represented as discrete motion tokens decoded into joint positions $J \in \mathbb{R}^{23 \times 3}$. 
Both past narration and future descriptions $y \in \{y^{\text{past}}, y^{\text{fut}}\}$ are natural language description of the corresponding motion segment. 
Our Ego3DLM produces a structured output sequence in a single autoregressive pass:
$(x^{\text{past}},\, x^{\text{fut}},\, y^{\text{past}},\, y^{\text{fut}})=f_{\theta}(\mathbf{P},\, \mathbf{V},\, \mathbf{S})$, where $f_\theta$ is the language model.

\noindent\textbf{Pose Representation and Tokenization.}
Following EgoLM~\cite{hong2025egolm}, human motion is modeled as a 23-joint kinematic tree with 6D rotation parameterization. Pose of each frame $t$ is represented as $M_t=[V_t^r, R_t^r, R_t^{rv}, R_t^j, R_t^{jv}]\in\mathbb{R}^{279}$, from which joint positions $J\in\mathbb{R}^{23\times3}$ are recovered via forward kinematics. 
To enable discrete generation, we train a PQ-VAE that maps pose sequences to indices over $C=4096$ codebooks, extending the LM's vocabulary with pose tokens $\{x\}_{c=1}^{C}$ and allowing the model to reason over pose and language in a unified token space.
Further details are found in the supplementary materials.

\noindent\textbf{Language Model.}
We build on a GPT-style causal LM (GPT-2, Qwen 2.5)~\cite{radford2019language, bai2025qwen2} trained primarily with next-token prediction. The LM comprises (i) a token embedding table (codebook), (ii) a Transformer~\cite{vaswani2017attention} backbone, and (iii) an output projection layer that maps hidden states to vocabulary logits. Non-text inputs $(\textbf{P},\textbf{V},\textbf{S})$ are injected as conditioning embeddings, enabling the model to exploit external context while preserving its linguistic priors. For motion tracking and forecasting, we use tokenized pose sequences, enabling the LM to generate motion outputs as discrete tokens. 

\subsection{3D Scene Feature Extraction}
\label{sec:scene}
We design a pipeline that injects semantic information from egocentric video into the 3D scene point cloud. Following 3D-LLM~\cite{hong20233d}, we generate dense frame-wise 2D semantic features using Mask2Former~\cite{cheng2022maskedattentionmasktransformeruniversal} instance masks and EVA-ViT-G~\cite{EVA} to fuse global and object-centric features. These are lifted onto the 3D point cloud using per-frame camera poses and a best-view selection policy, retaining features from the closest camera viewpoint for each 3D point to mitigate occlusion errors.
The point cloud is transformed into the ego initial frame for alignment, voxelized for efficiency, and encoded with egocentric 3D sinusoidal positional encodings. A Q-Former~\cite{li2023blip  } then compresses the scene into K=32 query embeddings. In parallel, per-frame CLIP~\cite{radford2021learning} image embeddings are linearly projected into the LM embedding space. The video and 3D scene query embeddings together form the final scene conditioning for Ego3DLM.

\begin{figure}
    \centering
    \includegraphics[width=0.99\linewidth]{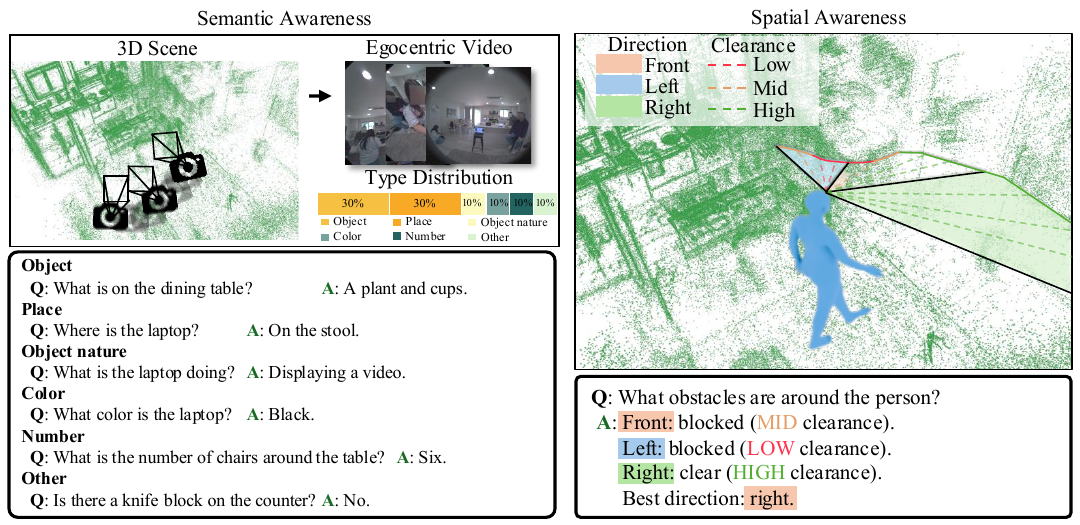} 
    \caption{
    \textbf{Spatial-semantic scene awareness QA dataset generation.} \textbf{Semantic Awareness:} For each 3D scene, we use egocentric videos to automatically generate question-answer pairs that describe and reason about the environment. We define six types of QA pairs targeting semantic understanding, and show their frequency distribution across the constructed data (left).
    \textbf{Spatial Awareness:} We partition the environment into three angular sectors (front, left, and right) and generate QA pairs to predict directional clearance levels and the most navigable direction (right).
    }
    \label{fig:qa_generate}
\end{figure}

\subsection{Stage I: Spatial-Semantic Scene Awareness Pretraining}
\label{sec:pretrain}
Human behavior is fundamentally constrained by environmental structure; therefore, understanding objects, their relations, and navigable free space is essential for downstream reasoning about human activity.
In this light, we initialize the language model with strong scene understanding by training it to reason over the spatial layout and semantic attributes of the surrounding 3D environment.
Given 3D scene embeddings $\mathbf{S}$ and egocentric video embeddings $\mathbf{V}$, we construct a dataset for spatial-semantic scene awareness by parsing the inputs into structured queries and answers as in Fig.~\ref{fig:qa_generate}. We then train the LM to jointly interpret semantic scene content (\eg objects, attributes, and places) and spatial structure (\eg obstacle layout and directional clearance).
This stage equips the LM with a spatial-semantic prior over environments before introducing human motion reasoning in later stages.

\noindent\textbf{Spatial Awareness ($\mathcal{Q}^{\text{spa}}$).}
We partition the environment into three angular sectors (front, left, and right) and train the model to predict directional clearance levels (low, mid, high), obstruction state (blocked, clear), and the most navigable direction (front, left, right), encouraging the LM to internalize obstacle layouts from egocentric observations.
\noindent\textbf{Semantic Scene Awareness ($\mathcal{Q}^{\text{sem}}$).}
We automatically generate scene-centric QA pairs capturing object identity, place relations, attributes, color, and counting using VLM~\cite{bai2025qwen2}, augmenting Nymeria~\cite{ma2024nymeria} which lacks explicit scene descriptions. We filter samples relying on unreliably reconstructed structures and remove human-centric attributes.

Overall, the resulting dataset contains approximately 535K spatial QA pairs and 115K semantic QA pairs across 208 scenes.
Further details on the dataset construction process for both are found in the supplementary materials.
Given $\mathbf{E}=(\mathbf{S}, \mathbf{V})$ and a semantic query $q$, the LM predicts the answer sequence $a$:
\begin{equation}
\mathcal{L}_{\text{Pre}} = -\sum_{i}\Bigg[
\underbrace{\log p_\theta\!\left(a_i^{\text{sem}} \mid a_{<i}^{\text{sem}}, q^{\text{sem}}, \mathbf{E}\right)}_{\mathcal{Q}^{\text{sem}}}
+
\underbrace{\log p_\theta\!\left(a_i^{\text{spa}} \mid a_{<i}^{\text{spa}}, q^{\text{spa}}, \mathbf{E}\right)}_{\mathcal{Q}^{\text{spa}}}
\Bigg],
\end{equation}
where semantic awareness queries and answers are denoted as $(q^{\text{sem}}, a^{\text{sem}})$ and spatial awareness queries and answers as $(q^{\text{spa}}, a^{\text{spa}})$, respectively.
This encourages the LM to internalize both semantic scene knowledge and spatial layout reasoning. This spatial-semantic prior forms the foundation for subsequent stages that require reasoning about human behavior within the environment.

\subsection{Stage II: Multi-Modal Multi-Task Instruction Tuning}
\label{sec:tuning}
We train the model using a unified instruction-tuning objective that jointly supervises four outputs: past motion, past motion description, future motion, and future motion description. Rather than predicting these outputs independently, the model generates them in a single autoregressive sequence, enabling shared reasoning across time and modalities.
A key design choice is to prepend an explicit \emph{spatial scene reasoning} step before any motion or language output. This encourages the model to first analyze spatial constraints in the environment, such as obstacles, free space, and navigable directions, before committing to predictions. The resulting generation process follows a chain-of-thought-like structure~\cite{ho2023large}, where spatial reasoning informs motion prediction, and motion prediction subsequently informs semantic description.

Conditioned on 3D scene embeddings $\textbf{S}$, egocentric video embeddings $\textbf{V}$, and three-point tracking features $\textbf{P}$, the model produces outputs in the following structured order: \texttt{[spatial scene description]}$\rightarrow$\texttt{[past \& future motion]} 
$\rightarrow$ \texttt{[past \& future motion description]}.
By generating a spatial scene description, motions, and motion descriptions in a single autoregressive pass, the model propagates structured context across the full output sequence, promoting three complementary forms of reasoning: \textit{spatial understanding}, where the model infers scene constraints to guide subsequent motion prediction; \textit{temporal understanding}, where past and future motion are jointly modeled so that observed trajectories directly inform future prediction; and \textit{semantic grounding}, where simultaneously predicting motions and descriptions enforces that the physical realization of a motion and its semantic description remain mutually consistent.
Given input $(\textbf{S}, \textbf{V}, \textbf{P})$ and a spatial query $q^{\text{spa}}$, we construct an instruction prompt $\textbf{I}$, as detailed in the supplementary materials. Given targets $(a^{\text{spa}}, x^{\text{past}}, x^{\text{fut}}, y^{\text{past}}, y^{\text{fut}})$, 
we optimize a unified objective that jointly supervises all output components:
\begin{equation}
\begin{aligned}
\mathcal{L}_{\text{IT}}
&=
\log p_\theta
\left(
a_i^{\text{spa}} \mid a_{<i}^{\text{spa}},\,  \textbf{I}
\right)
+
\log p_\theta
\left(
x_i^{\text{past}} \mid x_{<i}^{\text{past}},\, a^{\text{spa}},\, \mathbf{I}
\right)\\
&+
\log p_\theta
\left(
x_i^{\text{fut}} \mid x_{<i}^{\text{fut}},\, x^{\text{past}},\, a^{\text{spa}},\, \mathbf{I}
\right)
+
\log p_\theta
\left(
y_i^{\text{past}} \mid y_{<i}^{\text{past}},\, x^{\text{fut}},\, x^{\text{past}},\, a^{\text{spa}},\, \mathbf{I}
\right)\\
&+
\log p_\theta
\left(
y_i^{\text{fut}} \mid y_{<i}^{\text{fut}},\, y^{\text{past}},\, x^{\text{fut}},\, x^{\text{past}},\, a^{\text{spa}},\, \mathbf{I}
\right)\\
\end{aligned}
\end{equation}

\subsection{Stage III: Multi-Modal Reward GRPO}
\label{sec:grpo}
Supervised instruction tuning optimizes the likelihood of ground-truth outputs but does not directly encourage cross-modal coherence between simultaneously generated motions and descriptions. To address this, we introduce a reinforcement learning stage using Group Relative Policy Optimization (GRPO)~\cite{shao2024deepseekmath}, which optimizes the model toward outputs that are simultaneously accurate and internally consistent across modalities.

\noindent\textbf{GRPO Objective.}
For each prompt $\textbf{I}$, GRPO samples a group of $G$ candidate outputs $\{o_i\}_{i=1}^{G}$ from the old policy $\pi_{\theta_{\text{old}}}$, assigns each a scalar reward $r_i$, and updates the policy by maximizing:
\begin{equation}
\label{eq:grpo}
\begin{split}
\mathcal{J}_{\text{GRPO}}(\theta) =
\mathbb{E}\;\frac{1}{G}\sum_{i=1}^{G}\frac{1}{|o_i|}\sum_{t=1}^{|o_i|}
\Biggl\{
&\min\!\left[
\rho_{i,t}\,\hat{A}_{i,t},\;
\operatorname{clip}\!\left(\rho_{i,t},\, 1{-}\varepsilon,\, 1{+}\varepsilon\right)\hat{A}_{i,t}
\right] \\
&- \beta\,\mathbb{D}_{\mathrm{KL}}\!\left[\pi_\theta \,\|\, \pi_{\mathrm{ref}}\right]
\Biggr\},
\end{split}
\end{equation}
where $\rho_{i,t} = \pi_\theta(o_{i,t}|o_{i,<t},\textbf{I})\,/\,\pi_{\theta_{\mathrm{ref}}}(o_{i,t}|o_{i,<t},\textbf{I})$ is the importance ratio and the group-relative advantage is:
\begin{equation}
\hat{A}_{i,t} = \frac{r_i - \mathrm{mean}(\mathbf{r})}{\mathrm{std}(\mathbf{r})}, \quad \mathbf{r} = \{r_1, \ldots, r_G\}.
\end{equation}
This eliminates the need for a separate value network, as outputs scoring above the group mean are reinforced purely through within-group reward comparisons.

\noindent\textbf{Multi-Modal Reward.}
The scalar reward $r_i$ decomposes into intra-modal and inter-modal signals, applied symmetrically over both past and future outputs:
\begin{equation}
\label{eq:reward}
r = w_{m} \cdot \underbrace{\max(0,\, 1 - \mathrm{JPE})}_{R_{\mathrm{motion}}}
  + w_{t} \cdot \underbrace{\mathrm{BLEU}}_{R_{\mathrm{text}}}
  + w_{d} \cdot R_{\mathrm{matching}}
  + R_{\mathrm{format}}.
\end{equation}
$\mathrm{JPE}$ (global Joint Position Error) and $\mathrm{BLEU}$-4 are intra-modal rewards that independently measure motion accuracy and description quality against ground truth. $R_{\mathrm{format}}$ penalizes outputs that violate the structured output template. $w$ is the weighting constant between rewards.

The inter-modal reward $R_{\mathrm{matching}}$ is the key signal that supervised training cannot provide, directly measuring fidelity between the simultaneously predicted motion and description:
\begin{equation}
R_{\mathrm{matching}} = -\Big[
  d_{gp}\!\left(\mathbf{e}_{t_{\mathrm{gt}}},\, \mathbf{e}_{m_{\mathrm{pred}}}\right)
  + d_{pg}\!\left(\mathbf{e}_{t_{\mathrm{pred}}},\, \mathbf{e}_{m_{\mathrm{gt}}}\right)
  + d_{pp}\!\left(\mathbf{e}_{t_{\mathrm{pred}}},\, \mathbf{e}_{m_{\mathrm{pred}}}\right)
\Big],
\end{equation}
where $d(\cdot,\cdot)$ is Euclidean distance in a shared embedding space projected by modality-specific encoders~\cite{guo2022generating}, and $\mathbf{e}_{t}$, $\mathbf{e}_{m}$ denote text and motion embeddings respectively. 
Each term penalizes inconsistency between: (i) ground-truth description and predicted motion ($d_{gp}$), (ii) predicted description and ground-truth motion ($d_{pg}$), and (iii) the simultaneously predicted motion-description pair ($d_{pp}$). Together, they encourage the model to produce motion and description that are mutually grounded, a signal that likelihood-based training alone cannot provide.

\section{Experiments}

\subsection{Experimental Setup}
\noindent\textbf{Dataset.}
We conduct experiments on the Nymeria dataset~\cite{ma2024nymeria}, a large-scale in-the-wild egocentric human motion corpus recorded with Project Aria devices~\cite{engel2023project}, containing language narrations, egocentric video, and 3D point clouds.
We parse each recording into narrated motion segments and form past-future pairs by linking successive 5-second segments at 10 FPS.
The earlier segment serves as past context, and the subsequent segment as the future target. 
We split the data at the scene level so that training and validation sets do not share environments, and we stratify scenes by dominant motion pattern (largely static vs.\ dynamic human motion) to maintain a similar static/dynamic ratio in both splits.
Further details are found in the supplementary materials.

\noindent\textbf{Baselines.}
We compare ours with 3 baseline methods: EgoLM~\cite{hong2025egolm}, FIction~\cite{ashutosh2025fiction}, and UniEgoMotion~\cite{patel2025uniegomotion}, all of which predict future pose sequences.
EgoLM separately generates text narration for past motion tracking along with human poses.
Other baselines only generate motion without language.
All baseline methods utilize three-point tracking and egocentric video, and FIction and UniEgoMotion additionally utilize a 3D scene.
Results for EgoLM with a 3D scene can be found in the supplementary materials.

\noindent\textbf{Metrics.}
APE, JPE, and ADE are pose-related metrics that respectively measure the local joint pose accuracy, global joint pose accuracy, and global head pose accuracy.
$2s$ denotes the accuracy of the first 2 seconds.
For tracking we additionally use Upper and Lower for upper body and lower body APE, along with joint angle (J. A.) and Root. Joint angle (J.A.) measures the mean angular error across all body joints, and Root measures the positional error of the root joint which captures global translation accuracy.
For text, we use Bleu~\cite{papineni2002bleu}, $\text{Rouge}_L$~\cite{lin2004rouge}, SBert score~\cite{reimers2019sentence}, and R-precision.
\textit{R@N} denotes top-N R-precision accuracy of narration.
Motion-description matching distance metrics $d_{gp}$, $d_{pg}$, and $d_{pp}$ are defined as L2 distances in the embedding space produced by modality specific encoders~\cite{guo2022generating}, computed between ground truth description and predicted motion ($d_{gp}$), predicted description and ground truth motion ($d_{pg}$), and the predicted motion-description pair ($d_{pp}$), respectively.
$d_{pp}^{f}$ and $d_{pp}^{p}$ denote the motion-description matching distances for the future and past.
For pretraining on $\mathcal{Q}^{spa}$, collision (obstruction) and freespace metrics measure the mean accuracy of all directions.
Further details are found in the supplementary materials.

\noindent\textbf{Implementation Details.} 
We use a GPT-2 medium as the language model backbone.
Full training settings, additional details, including pretraining and instruction templates, 3D scene QA dataset generation, and additional results with Qwen 2.5 backbone, are provided in the supplementary materials. 

\begin{table}[t]
\centering
\caption{Comparison of prediction and tracking results on motion and text metrics. $x$-$y$ Align. denotes motion and text alignment. Embedding distance functions ($d_{gp}, d_{pg}, d_{pp}$) are defined in Sec.~\ref{sec:grpo}. Best are denoted in \textbf{bold}. Arrow denotes better.}
\small

\resizebox{\textwidth}{!}{
\begin{tabular}{lccccccccccccc}
\toprule
\multirow{2}{*}{Method} & \multicolumn{8}{c}{Motion Prediction (3 modes)} & \multicolumn{5}{c}{Motion Tracking} \\ \cmidrule(lr){2-9} \cmidrule(lr){10-14}
 & APE $\downarrow$ & JPE $\downarrow$ & ADE $\downarrow$ & ADE$_{2s}$ $\downarrow$ & FDE $\downarrow$ & FDE$_{2s}$ $\downarrow$ & FID $\downarrow$ & Diversity $\uparrow$ & APE $\downarrow$ & Upper $\downarrow$ & Lower $\downarrow$ & J.A. $\downarrow$ & Root $\downarrow$ \\ \midrule
Fiction~\cite{ashutosh2025fiction} & 206.2 & 564.7 & 558.3 & 416.7 & 904.3 & 494.9 & 0.3275 & 0.7432 & 181.1 & 114.0 & 282.3 & 31.28 & 34.81 \\
EgoLM~\cite{hong2025egolm} (GT motion) & 168.8 & 583.9 & 540.0 & 299.0 & 1,059.5 & 494.8 & 1.3840 & 0.0055 & - & - & - & - & - \\
EgoLM~\cite{hong2025egolm} (Inst. tuning) & 184.9 & 579.4 & 552.6 & 329.9 & 983.6 & 519.2 & 0.2137 & 0.6142 & 161.9 & 95.6 & 265.6 & 33.63 & 24.01 \\
UniEgoMotion~\cite{patel2025uniegomotion} & 151.5 & 424.3 & 409.7 & 223.9 & 720.5 & 360.4 & 0.1530 & 0.9022 & 152.2 & 79.4 & 233.3 & 26.48 & 22.71 \\
\textbf{Ours (Ego3DLM)} & \textbf{147.9} & \textbf{364.5} & \textbf{343.9} & \textbf{205.9} & \textbf{648.1} & \textbf{312.6} & \textbf{0.0160} & \textbf{1.0624} & \textbf{96.4} & \textbf{53.1} & \textbf{152.7} & \textbf{22.30} & \textbf{19.57} \\ \bottomrule
\end{tabular}
}

\vspace{0.4em}
\resizebox{\textwidth}{!}{
\begin{tabular}{lcccccc cccccc c}
\toprule
\multirow{2}{*}{Method} & \multicolumn{6}{c}{Future Motion Description} & \multicolumn{6}{c}{Past Motion Narration} & \multicolumn{1}{c}{$x$-$y$ Align.} \\ \cmidrule(lr){2-7} \cmidrule(lr){8-13} \cmidrule(lr){14-14}
 & Bleu-4 $\uparrow$ & Bleu-1 $\uparrow$ & Rogue-L $\uparrow$ & SBert $\uparrow$ & R@3 $\uparrow$ & $d_{pg}$ $\downarrow$ & Bleu-4 $\uparrow$ & Bleu-1 $\uparrow$ & Rogue-L $\uparrow$ & SBert $\uparrow$ & R@3 $\uparrow$ & $d_{pg}$ $\downarrow$ & $d_{pp}$ $\downarrow$ \\ \midrule
LLM (Qwen 2.5 7B~\cite{bai2025qwen2}) & 0.0202 & 0.0970 & 0.2343 & 0.6120 & 0.1940 & 6.6730 & 0.0232 & 0.1477 & 0.2267 & 0.6183 & 0.1944 & 6.9648 & - \\
EgoLM~\cite{hong2025egolm} (Inst. tuning) & - & - & - & - & - & - & 0.0649 & 0.3002 & 0.2642 & 0.5900 & 0.2639 & 6.8992 & 9.8686 \\
\textbf{Ours (Ego3DLM)} & \textbf{0.1039} & \textbf{0.3839} & \textbf{0.3004} & \textbf{0.6206} & \textbf{0.2911} & \textbf{6.3507} & \textbf{0.1107} & \textbf{0.3966} & \textbf{0.3195} & \textbf{0.6458} & \textbf{0.4264} & \textbf{4.9470} & \textbf{4.2571} \\ \bottomrule
\end{tabular}
}
\label{tab:main}
\end{table}

\subsection{Quantitative Prediction}
As shown in Table~\ref{tab:main}, Ego3DLM achieves state-of-the-art performance across all motion prediction and tracking metrics.
For prediction, our model outperforms the strongest baseline UniEgoMotion across all metrics, reducing APE by $2.4\%$, JPE by $14.1\%$, ADE$_{2s}$ by $8.0\%$, and FDE$_{2s}$ by $13.3\%$, while also achieving the best FID and highest diversity, indicating that our predictions are both accurate and diverse.
The improvement is even more pronounced for motion tracking, where Ego3DLM reduces APE by $36.7\%$ and upper/lower body APE by $33.1\%$/$34.5\%$ over UniEgoMotion, demonstrating that joint past-and-future reasoning with explicit 3D scene conditioning yields a substantially more accurate understanding of observed motion.

\begin{figure*}[!t]
    \centering
    \includegraphics[width=1.0\linewidth]{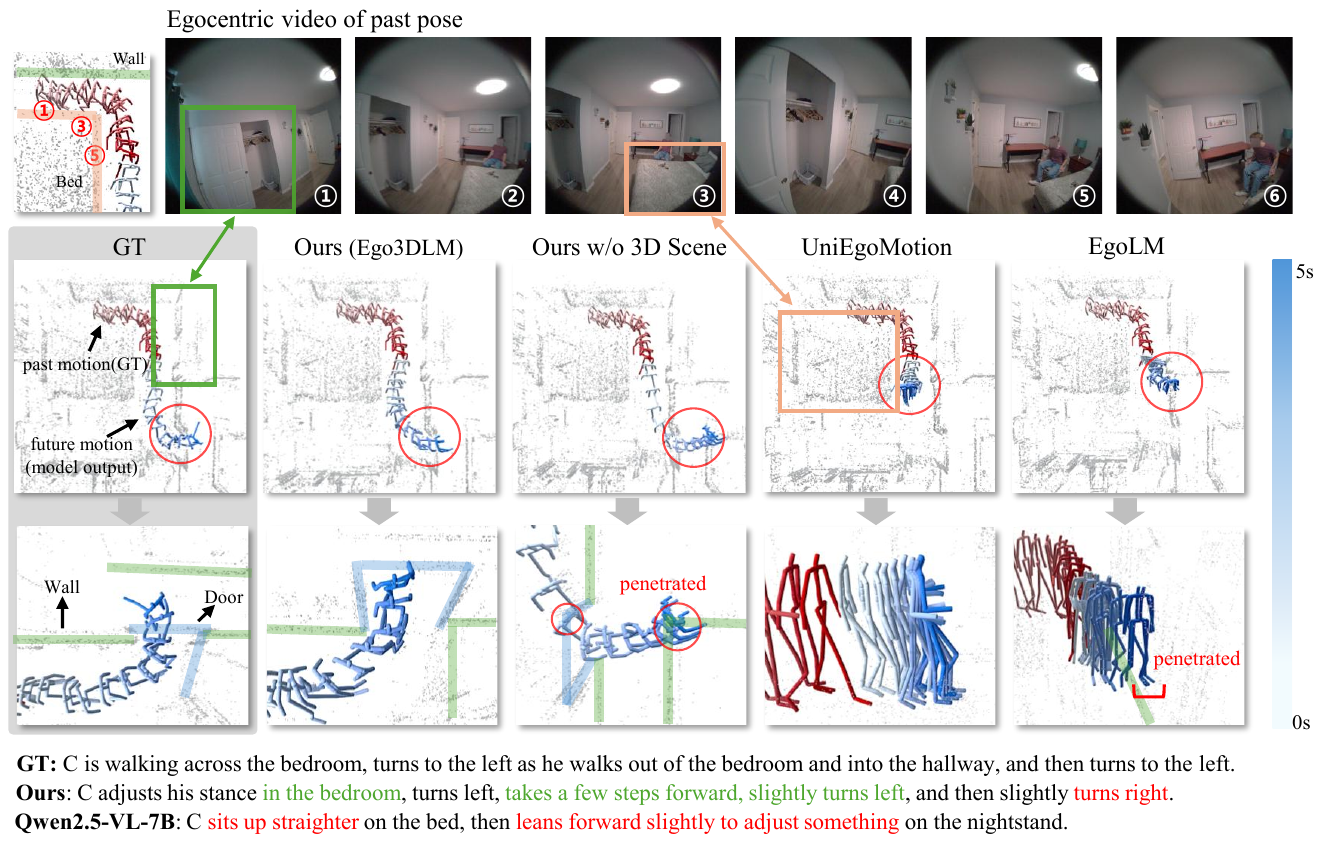}
    \caption{
        Qualitative results for motion forecasting and tracking.
        The ground-truth (GT) past motion trajectory is shown in \textcolor{red}{\textbf{red}}, and the predicted future motion is visualized in \textcolor{blue}{\textbf{blue}}.
        For future motion narration below the figure, \textcolor{darkgreen}{\textbf{green}} highlights correct phrases matching the GT, while \textcolor{red}{\textbf{red}} highlights incorrect or mismatched descriptions.
    }
    \label{fig:fig_qual}
\end{figure*}

One aspect to note is that Ego3DLM's holistic prediction is superior to modality-specific or separately generated outputs.
Methods that predict only motion (FIction, UniEgoMotion) forgo language entirely, while a general-purpose LLM (Qwen 2.5 7B) produces text without any motion grounding, and EgoLM, despite generating both modalities, does so through separate decoding streams.
In contrast, our single-pass joint decoding achieves substantially higher scores across all text metrics, improving future motion description Bleu-4 by $60.0\%$ and past motion narration Bleu-4 by $70.4\%$ over EgoLM.
Most strikingly, the motion-description embedding distance ($x$-$y$ Align.) improves by $56.8\%$, indicating that our simultaneously predicted motion and descriptions are tightly aligned in a shared semantic space.
This demonstrates that Ego3DLM produces physically plausible motions whose simultaneously generated motion descriptions faithfully reflect the underlying motion dynamics, confirming that holistic joint prediction yields semantically coherent motion-language outputs.

\subsection{Qualitative Prediction}
Figure~\ref{fig:fig_qual} presents qualitative comparisons among UniEgoMotion, EgoLM, and our Ego3DLM with and without 3D scene. 
For motion forecasting, ours without 3D scene fails to incorporate the room layout and collides with a wall.
While UniEgoMotion leverages 3D scene to avoid collision, it generates a contextually irrelevant motion that abruptly stops.
As EgoLM does not leverage 3D scene, the predicted motion results in collision.
Our method generates an accurate sequence that predicts walking motion around the bed to the door.
Notably, the future motion description accurately depicts the predicted motion, demonstrating a comprehensive knowledge of human motion intent in both motion and language domain.
Overall, our model produces motions that best conform to both physical constraints and past motion context, outperforming baselines with or without 3D spatial context.

\subsection{Ablation Study}
\subsubsection{Pretraining.}
Table~\ref{tab:ablation_pretraining} ablates the effect of input modality and pretraining target on scene understanding.
First, combining both video and 3D scene inputs consistently improves performance over video alone across most metrics for both semantic awareness $\mathcal{Q}^{sem}$ and spatial awareness $\mathcal{Q}^{spa}$, confirming that 3D scene features provide complementary spatial information beyond what egocentric video alone can convey.
Second, jointly training on both spatial and semantic QA objectives yields the best results across all metrics, with Bleu-1, SBert, and Rouge$_L$ all reaching their peak while simultaneously achieving the highest Collision and Freespace scores.
Notably, adding spatial awareness improves semantic understanding beyond training on semantic awareness alone, and vice versa, suggesting that the two objectives are mutually reinforcing rather than competing.
This indicates that reasoning about spatial layout and semantic scene content are complementary skills, and that exposure to both during pretraining allows Ego3DLM to acquire a thorough understanding of the 3D environment.

\begin{table}[t]
\centering
\caption{Ablation on pretraining. The effects of input modality (video, 3D scene) and pretraining objective ($\mathcal{Q}^{sem}$, $\mathcal{Q}^{spa}$) are analyzed.}
\small
\setlength{\tabcolsep}{4pt}

\resizebox{\textwidth}{!}{
\begin{tabular}{>{\centering\arraybackslash}p{2.5cm}cc @{\hspace{20pt}} cccccc}
\toprule
Objective & Video & 3D Scene & Bleu-1$\uparrow$ & Bleu-4$\uparrow$ & SBert$\uparrow$ & Rouge-L$\uparrow$ & $\text{Collision}_{\text{mean}}\uparrow$ & $\text{Freespace}_{\text{mean}}\uparrow$ \\ \midrule
\multirow{2}{*}{$\mathcal{Q}^{\text{sem}}$} & \checkmark & & 0.5298 & 0.2116 & 0.7403 & 0.5244 & - & - \\
 & \checkmark & \checkmark & 0.5330 & \textbf{0.2478} & 0.7393 & 0.5219 & - & - \\ \midrule
\multirow{2}{*}{$\mathcal{Q}^{\text{spa}}$} & \checkmark & & - & - & - & - & 0.7387 & 0.8720 \\
 & \checkmark & \checkmark & - & - & - & - & 0.7428 & 0.8728 \\ \midrule
$\mathcal{Q}^{\text{sem}}$ + $\mathcal{Q}^{\text{spa}}$ & \checkmark & \checkmark & \textbf{0.5399} & 0.2469 & \textbf{0.7438} & \textbf{0.5278} & \textbf{0.7764} & \textbf{0.8848} \\ \bottomrule
\end{tabular}
}
\label{tab:ablation_pretraining}
\end{table}

\begin{table*}[t]
\centering
\caption{Ablation on instruction tuning. The effects of pretraining ($\mathcal{Q}^{sem}$, $\mathcal{Q}^{spa}$), use of spatial scene reasoning (SSR) and 3D scene ($\textbf{S}$) are compared.}
\label{tab:ablation_instruction}

\resizebox{\linewidth}{!}{
\begin{tabular}{l cccccc cccc}
\toprule
\multirow{2}{*}{Method} & \multicolumn{6}{c}{Motion Prediction (3 modes)} & \multicolumn{4}{c}{Future Motion Description} \\
\cmidrule(lr){2-7} \cmidrule(lr){8-11}
 & APE $\downarrow$ & JPE $\downarrow$ & ADE$_{2s}$ $\downarrow$ & FDE$_{2s}$ $\downarrow$ & FID $\downarrow$ & Diversity & R@3 $\uparrow$ & R@2 $\uparrow$ & R@1 $\uparrow$ & $d_{pg}$ $\downarrow$ \\ \midrule
 Ours SFT only & \textbf{148.5} & \textbf{368.3} & \textbf{208.8} & \textbf{317.9} & \textbf{0.0190} & 0.9358 & \textbf{0.2706} & \textbf{0.1903} & \textbf{0.1072} & 6.4837 \\ \midrule
w/o $\mathcal{Q}^{sem}$   & 153.5 & 381.1 & 221.9 & 335.9 & 0.0336 & 0.9644 & 0.2558 & 0.1760 & 0.0987 & 6.5460 \\
w/o $\mathcal{Q}^{spa}$   & 152.9 & 378.3 & 214.3 & 329.5 & 0.0335 & 1.0122 & 0.2453 & 0.1725 & 0.0894 & 6.7133 \\
w/o SSR       & 151.5 & 376.6 & 213.4 & 325.3 & 0.0231 & \textbf{1.0830} & 0.2638 & 0.1859 & 0.1021 & \textbf{6.4494} \\
w/o $\textbf{S}$  & 156.2 & 388.5 & 220.8 & 334.3 & 0.0426 & 0.9352 & 0.2574 & 0.1809 & 0.0945 & 6.6789 \\
\bottomrule
\end{tabular}
}

\vspace{0.4em}

\resizebox{\linewidth}{!}{
\begin{tabular}{l ccccc cccc c}
\toprule
\multirow{2}{*}{Method} & \multicolumn{5}{c}{Motion Tracking} & \multicolumn{4}{c}{Past Motion Narration} & \multicolumn{1}{c}{$x$-$y$ Align.} \\
\cmidrule(lr){2-6} \cmidrule(lr){7-10} \cmidrule(lr){11-11}
 & APE $\downarrow$ & Upper $\downarrow$ & Lower $\downarrow$ & J.A. $\downarrow$ & Root $\downarrow$ & \makecell{R@3 $\uparrow$} & \makecell{R@2 $\uparrow$} & \makecell{R@1 $\uparrow$} & \makecell{$d_{pg}$ $\downarrow$} & $d_{pp}$ $\downarrow$ \\ \midrule
 Ours SFT only & \textbf{98.3} & 53.4 & \textbf{155.3} & \textbf{22.52} & 19.62 & \textbf{0.4157} & 0.3022 & 0.1693 & \textbf{5.0049} & \textbf{5.9170} \\ \midrule
w/o $\mathcal{Q}^{sem}$   & 122.1 & 69.6 & 190.1 & 27.84 & 25.40 & 0.3960 & 0.2961 & 0.1674 & 5.1349 & 5.9190 \\
w/o $\mathcal{Q}^{spa}$   & 99.3 & 53.6 & 158.7 & 22.79 & 19.93 & 0.3987 & 0.2971 & \textbf{0.1697} & 5.2278 & 6.1612 \\
w/o SSR       & 99.4 & 53.1 & 159.8 & 22.64 & \textbf{19.25} & 0.4090 & \textbf{0.3069} & 0.1661 & 5.0103 & 6.3930 \\
w/o $\textbf{S}$  & 105.8 & 57.5 & 168.8 & 24.03 & 20.92 & 0.3955 & 0.2837 & 0.1566 & 5.2103 & 6.3012 \\
\bottomrule
\end{tabular}
}
\end{table*}

\begin{table}[t]
\centering
\caption{Ablation on GRPO finetuning. A more comprehensive reward signal results in overall improvement, especially the semantic understanding of past and future motion. $d_{pp}^f$ and $d_{pp}^p$ each denote the embedding distance between generated motion and description for future and past, respectively.}
\vspace{-0.8em} 
\footnotesize 
\setlength{\tabcolsep}{2.5pt} 
\renewcommand{\arraystretch}{1.05}

\resizebox{\textwidth}{!}{
\begin{tabular}{ccc @{\hspace{12pt}} ccccc ccccc}
\toprule
\multicolumn{3}{c}{Reward} & \multicolumn{5}{c}{Motion Prediction (3 modes)} & \multicolumn{5}{c}{Motion Tracking} \\ \cmidrule(lr){1-3} \cmidrule(lr){4-8} \cmidrule(lr){9-13}
{\scriptsize $R_{\text{motion}}$} & {\scriptsize $R_{\text{text}}$} & {\scriptsize $R_{\text{matching}}$} & {\scriptsize APE $\downarrow$} & {\scriptsize JPE $\downarrow$} & {\scriptsize ADE$_{2s}$ $\downarrow$} & {\scriptsize FDE$_{2s}$ $\downarrow$} & {\scriptsize FID} & {\scriptsize APE $\downarrow$} & {\scriptsize Upper $\downarrow$} & {\scriptsize Lower $\downarrow$} & {\scriptsize J.A. $\downarrow$} & {\scriptsize Root $\downarrow$} \\ \midrule
\multicolumn{3}{c}{SFT only} & 148.5 & 368.3 & 208.8 & 317.9 & 0.0190 & 98.3 & 53.4 & 155.3 & 22.52 & 19.62 \\ \midrule
\checkmark & & & 148.1 & \textbf{364.2} & \textbf{204.8} & \textbf{310.5} & \textbf{0.0149} & 96.6 & \textbf{53.1} & 153.2 & 22.43 & 19.80 \\
\checkmark & \checkmark & & 148.7 & 369.6 & 206.2 & 313.8 & 0.0177 & 97.4 & 53.4 & 153.6 & 22.55 & 19.60 \\
\checkmark & \checkmark & \checkmark & \textbf{147.9} & 364.5 & 205.9 & 312.6 & 0.0160 & \textbf{96.4} & \textbf{53.1} & \textbf{152.7} & \textbf{22.30} & \textbf{19.57} \\ \bottomrule
\label{table:ablation_grpo}
\end{tabular}
}

\vspace{0.2em}

\resizebox{\textwidth}{!}{
\begin{tabular}{ccc @{\hspace{12pt}} cccc cccc cc}
\toprule
\multicolumn{3}{c}{Reward} & \multicolumn{4}{c}{Future Motion Description} & \multicolumn{4}{c}{Past Motion Narration} & \multicolumn{2}{c}{$x$-$y$ Align.} \\ \cmidrule(lr){1-3} \cmidrule(lr){4-7} \cmidrule(lr){8-11} \cmidrule(lr){12-13}
{\scriptsize $R_{\text{motion}}$} & {\scriptsize $R_{\text{text}}$} & {\scriptsize $R_{\text{matching}}$} & {\scriptsize Bleu-4 $\uparrow$} & {\scriptsize Bleu-1 $\uparrow$} & {\scriptsize R@3 $\uparrow$} & {\scriptsize $d_{pg}$ $\downarrow$} & {\scriptsize Bleu-4 $\uparrow$} & {\scriptsize Bleu-1 $\uparrow$} & {\scriptsize R@3 $\uparrow$} & {\scriptsize $d_{pg}$ $\downarrow$} & {\scriptsize $d_{pp}^f$ $\downarrow$} & {\scriptsize $d_{pp}^p$ $\downarrow$} \\ \midrule
\multicolumn{3}{c}{SFT only} & 0.1036 & 0.3789 & 0.2706 & 6.4837 & 0.1126 & 0.3930 & 0.4157 & 5.0049 & 5.9170 & 4.2861 \\ \midrule
\checkmark & & & 0.1019 & 0.3716 & 0.2792 & 6.3974 & 0.1148 & 0.3936 & 0.4206 & \textbf{4.9424} & 5.9181 & 4.3260 \\
\checkmark & \checkmark & & 0.1037 & 0.3815 & 0.2726 & 6.4560 & \textbf{0.1160} & 0.3944 & 0.4071 & 5.0080 & 5.8519 & 4.3583 \\
\checkmark & \checkmark & \checkmark & \textbf{0.1039} & \textbf{0.3839} & \textbf{0.2911} & \textbf{6.3507} & 0.1107 & \textbf{0.3966} & \textbf{0.4264} & 4.9470 & \textbf{5.8063} & \textbf{4.2571} \\ \bottomrule
\end{tabular}
}
\end{table}

\subsubsection{Instruction Tuning.}
Table~\ref{tab:ablation_instruction} ablates the contribution of each component during instruction tuning across all four tasks.
Removing 3D scene input causes the most consistent and severe degradation across all tasks, confirming that explicit 3D scene conditioning is the single most important factor for accurate motion prediction, as the model cannot adequately infer physical constraints from egocentric video alone.
Removing semantic QA pretraining leads to a substantial drop in motion tracking performance, suggesting that semantic scene understanding provides crucial context for interpreting observed body motion relative to the environment.
Removing spatial QA similarly degrades both motion prediction and cross-modal alignment, indicating that spatial awareness of navigable directions and clearance is particularly important for generating physically plausible future motion and grounding them in language.
Together, these results corroborate the pretraining ablation findings: semantic and spatial understanding are complementary, and both contribute to downstream prediction quality.
Finally, removing spatial scene reasoning (SSR) leads to notable degradation in motion-description alignment and future motion description quality.
This demonstrates that the structured scene reasoning step does not merely reformat outputs; it actively elicits spatially grounded reasoning that propagates through the generation process, improving the semantic coherence between predicted motions and their descriptions.

\subsubsection{GRPO Finetuning.}
Table~\ref{table:ablation_grpo} validates the contribution of each reward component.
As expected, adding $R_{\text{motion}}$ primarily improves motion prediction (JPE by $1.1\%$, FID by $21.6\%$), and further adding $R_{\text{text}}$ improves narration quality (Bleu-4 by $3.0\%$), confirming that intra-modal rewards effectively sharpen their respective output modalities.
Notably, incorporating the inter-modal matching reward $R_{\text{matching}}$ yields broad improvements across \emph{both} motion and text metrics simultaneously, including motion prediction, narration Bleu scores, and motion-description alignment distances ($d^f_{pp}$ by $0.8\%$, $d^p_{pp}$ by $2.3\%$ over the previous stage).
This reciprocal improvement demonstrates that explicitly rewarding motion-description consistency encourages the model to develop a comprehensive and semantically grounded understanding of human motion intent. By enforcing predicted motions and descriptions to be mutually coherent, the model learns richer cross-modal representations that benefit each modality independently.

\section{Conclusion}
We present Ego3DLM, a unified 3D-aware framework for holistic tracking and forecasting of all four outputs in a single autoregressive pass: past motion, future motion, past narration, and future description. By leveraging three-point tracking, egocentric video, and explicit 3D scene features, Ego3DLM establishes a new state-of-the-art across all motion prediction, tracking, and language description benchmarks. This is achieved by spatial-semantic scene awareness pretraining, holistic multi-task instruction tuning, and multi-modal reward GRPO finetuning that directly optimizes motion-language fidelity. Our experiments demonstrate consistent improvements over baselines, particularly in the tight coupling of predicted motion and its description, marking a step toward more anticipatory, interpretable, and interactive embodied AI systems.

\noindent\textbf{Limitations and Future Works.}
Since most human activities take place in familiar venues where users spend substantial time and accumulate rich 3D spatial knowledge, we assume the 3D scene feature is given.
While reasonable for most XR applications, reliance on a precomputed 3D scene remains a limitation.
This opens up promising future work toward a language model that can incrementally register 3D knowledge of newly observed scenes on the fly and reason about how a human will interact within them.

\section*{Acknowledgements}
This work was supported by the Institute of Information \& communications Technology Planning \& Evaluation (IITP) grant funded by the Korea government(MSIT) (No. RS-2024-00457882, AI Research Hub Project), by Center for Advanced Urban Systems (CAUS) of Korea Advanced Institute of Science and Technology (KAIST) funded by GS E\&C, and by the InnoCORE program of the Ministry of Science and ICT(N10250156).

%
%
\bibliographystyle{splncs04}
\bibliography{main}

\clearpage
\setcounter{section}{0}
\setcounter{subsection}{0}
\setcounter{table}{0}
\setcounter{figure}{0}
\setcounter{equation}{0}
\renewcommand{\thesection}{\Alph{section}}
\renewcommand{\thetable}{\Alph{table}}
\renewcommand{\thefigure}{\Alph{figure}}
\renewcommand{\theequation}{\Alph{equation}}
\renewcommand{\thesubsection}{\thesection\arabic{subsection}}
\renewcommand{\theHsection}{S\arabic{section}}
\renewcommand{\theHsubsection}{S\arabic{section}.\arabic{subsection}}
\renewcommand{\theHtable}{S\arabic{table}}
\renewcommand{\theHfigure}{S\arabic{figure}}
\renewcommand{\theHequation}{S\arabic{equation}}

\begin{center}
    {\Large\bfseries Supplementary Materials for\\[4pt]
     Ego-Human Motion Prediction with 3D-Aware LLM\par}
\end{center}
\vspace{1.5em}

\section{Overview}
Our supplementary material provides further clarification of the proposed method and offers additional insights and in-depth discussions on areas not extensively covered in the main paper:
\begin{itemize}
    \item Corrections to Minor Errors in Main Paper (Sec.~\ref{sec:errata}).
    \item Implementation Details (Sec.~\ref{sec:impl_supp}).
    \item Additional Experimental Results (Sec.~\ref{sec:exp_supp}).
    \item Additional Qualitative Results (Sec.~\ref{sec:qual_supp}).
\end{itemize}

\section{Corrections to Minor Errors in Main Paper}
\label{sec:errata}
The caption of Fig.~\textcolor{red}{2} in the main paper contains a minor error. In particular, the description of the pretraining process (stage I) was inaccurate. Our pretraining learns spatial and semantic awareness in 3D scenes through two complementary objectives. We apologize for any confusion caused by this error.

\section{Implementation Details}
\label{sec:impl_supp}
\subsection{Model Architecture and Training Details}
We use a GPT-2 medium backbone for the language model as in EgoLM~\cite{hong2025egolm}. For motion tokenization, we use a product-quantized VAE (PQ-VAE) following EgoLM, employing a convolutional encoder–decoder architecture with a temporal downsampling rate of $4$, a codebook of $4096$ entries with $64$-dimensional codes, and exponential moving average updates with codebook reset~\cite{dhariwal2020jukebox}. We use two codebooks for the PQ-VAE. 
Motion sequences are sampled at 
$10$Hz and have lengths between $45$ and $300$ frames. For scenes, we retain maximum \(10{,}000\) points after voxelization and sampling. 
The Q-Former is initialized from pretrained BERT-base~\cite{devlin2019bert} and uses $32$ learned query tokens. For modality-specific projection layers, we use a linear layer for 3D scene embeddings, a small MLP for three-point tracking features, and another small MLP for video frame embeddings.
The three-point tracking features are constructed by concatenating the 3D translation, 6D rotation, 3D translation velocity, and 6D rotation velocity for all three points along the channel dimension, resulting in a 54-dimensional representation.

In both pre-training and instruction tuning, we train with AdamW~\cite{LoshchilovH19} optimizer (learning rate $1\times10^{-4}$) and a cosine-annealing learning rate scheduler~\cite{loshchilov2017sgdr}, using a batch size of $16$. 
%
%
During the multi-modal reward GRPO-based reinforcement finetuning, $G$ is set to $6$, and the reward weights are established at $w_m=1.0$, $w_t=0.8$, and $w_d=0.02$. The clipping parameter $\epsilon$ and KL coefficient $\beta$ are assigned values of $0.2$ and $0.001$, respectively. Regarding $R_{\text{format}}$, a score of 0 is granted if the generated output follows the structure described in Sec.~\ref{supp_sec:instruction_tuning}; otherwise, a penalty of up to $-3$ is applied. Finally, the learning rate and batch size are set to $1\times10^{-6}$ and $4$,  respectively.

\subsection{Pose Representation Details}
Human motion is modeled as a kinematic tree of body joints with a root joint that carries global translation and rotation over time. We adopt a kinematic tree comprising 23 joints and a 6D rotation representation for stable regression while capturing dynamics. 
At time $t$, the pose state comprises (i) root translational velocity $V_t^r\in\mathbb{R}^3$, (ii) root rotation $R_t^r\in\mathbb{R}^6$, (iii) root rotational velocity $R_t^{rv}\in\mathbb{R}^6$, (iv) joint rotations $R_t^j\in\mathbb{R}^{22\times6}$, and (v) joint rotational velocities $R_t^{jv}\in\mathbb{R}^{22\times6}$.
Concatenating these terms yields a vector $M_t=[V_t^r, R_t^r, R_t^{rv}, R_t^j, R_t^{jv}]\in\mathbb{R}^{279}$, and a sequence $M=\{M_t\}_{t=1}^T$ represents a motion clip. Joint positions $J\in\mathbb{R}^{23\times3}$ are recovered via forward kinematics from $\{M_t\}$.

\subsection{Pretraining Templates}
For spatial-semantic scene awareness LM pretraining, we design a pair of prompt templates for spatial and semantic reasoning to enhance scene understanding.
The exact templates are as follows. The 3D scene and video embeddings shared by all tasks are injected via $<$\textit{3D\_Scene\_Placeholder}$>$ and $<$\textit{Video\_Placeholder}$>$. The natural language query and response are placed at $<$\emph{Question\_Placeholder}$>$ and $<$\emph{Answer\_Placeholder}$>$, respectively. 
These placeholders standardize how \emph{scene and video embeddings} and \emph{text} are concatenated and fed into the LM, enabling consistent training across pretraining objectives.

\begin{tcolorbox}[colback=white, colframe=black, sharp corners, boxrule=0.5pt, fontupper=\small, breakable, left=3mm, 
right=3mm, 
]
    \textbf{Spatial Scene Awareness} \\
    \textbf{Input:} \textit{Answer the question with $<$3D\_Scene\_Placeholder$>$ scene and the corresponding $<$Video\_Placeholder$>$ video frames: What obstacles are around the person in the current frame?}\\
    \textbf{Output:} \textit{$<$Answer\_Placeholder$>$}
    
    \vspace{0.5cm}
    \textbf{Semantic Scene Awareness} \\
    \textbf{Input:} \textit{Answer the question with $<$3D\_Scene\_Placeholder$>$ scene and the corresponding $<$Video\_Placeholder$>$ video frames: $<$Question\_Placeholder$>$}\\
    \textbf{Output:} \textit{$<$Answer\_Placeholder$>$}
%
\end{tcolorbox}

\subsection{Instruction Tuning Templates}
\label{supp_sec:instruction_tuning}
The instruction template for a unified instruction tuning objective that jointly outputs motion tracking, motion forecasting, past motion description, and future motion description is shown below. The 3D scene embeddings and video embeddings are inserted into $<$\textit{3D\_Scene\_Placeholder}$>$ and $<$\textit{Video\_Placeholder}$>$, respectively. 
Similarly, the embeddings for three-point motion tracking are placed at $<$\emph{TP\_Placeholder}$>$.
The output targets, including the spatial reasoning answer, past motion, future motion, past motion description, and future motion description, are tokenized and inserted into their corresponding placeholders.
Delimiter tokens are used to structure prompts across tasks and modalities, but are omitted from the template below for simplicity.

\begin{tcolorbox}[colback=white, colframe=black, sharp corners, boxrule=0.5pt, fontupper=\small, breakable,
left=3mm, 
right=3mm, 
]
    \textbf{Multi-Modal Multi-Task Instruction Tuning} \\
    \textbf{Instruction:} 
    \textit{Perform human motion tracking and prediction. First, analyze the spatial constraints (front, left, right) and the optimal direction from the scene. Then, output current and future human motion token sequences, and describe current and future human motion conditioned on the given scene, observed video CLIP embeddings, and observed three-points features. \\
    Input scene: $<$3D\_Scene\_Placeholder$>$. \\
    Input video CLIP embeddings: $<$Video\_Placeholder$>$.\\
    Input three-points features: $<$TP\_Placeholder$>$}.\\
    \textbf{Output:} \textit{$<$Answer\_Placeholder$>$$<$Past\_Motion\_Placeholder$>$$<$Future\_Motion\_\\Placeholder$>$$<$Past\_Description\_Placeholder$>$$<$Future\_Description\_Placeholder$>$}
    %
\end{tcolorbox}

\subsection{3D Scene QA Dataset Generation}
%
\subsubsection{Semantic Awareness ($\mathcal{Q}_{\text{sem}}$) Dataset Generation.}
\begin{figure*}[!t]
    \centering
    \includegraphics[width=1\linewidth]{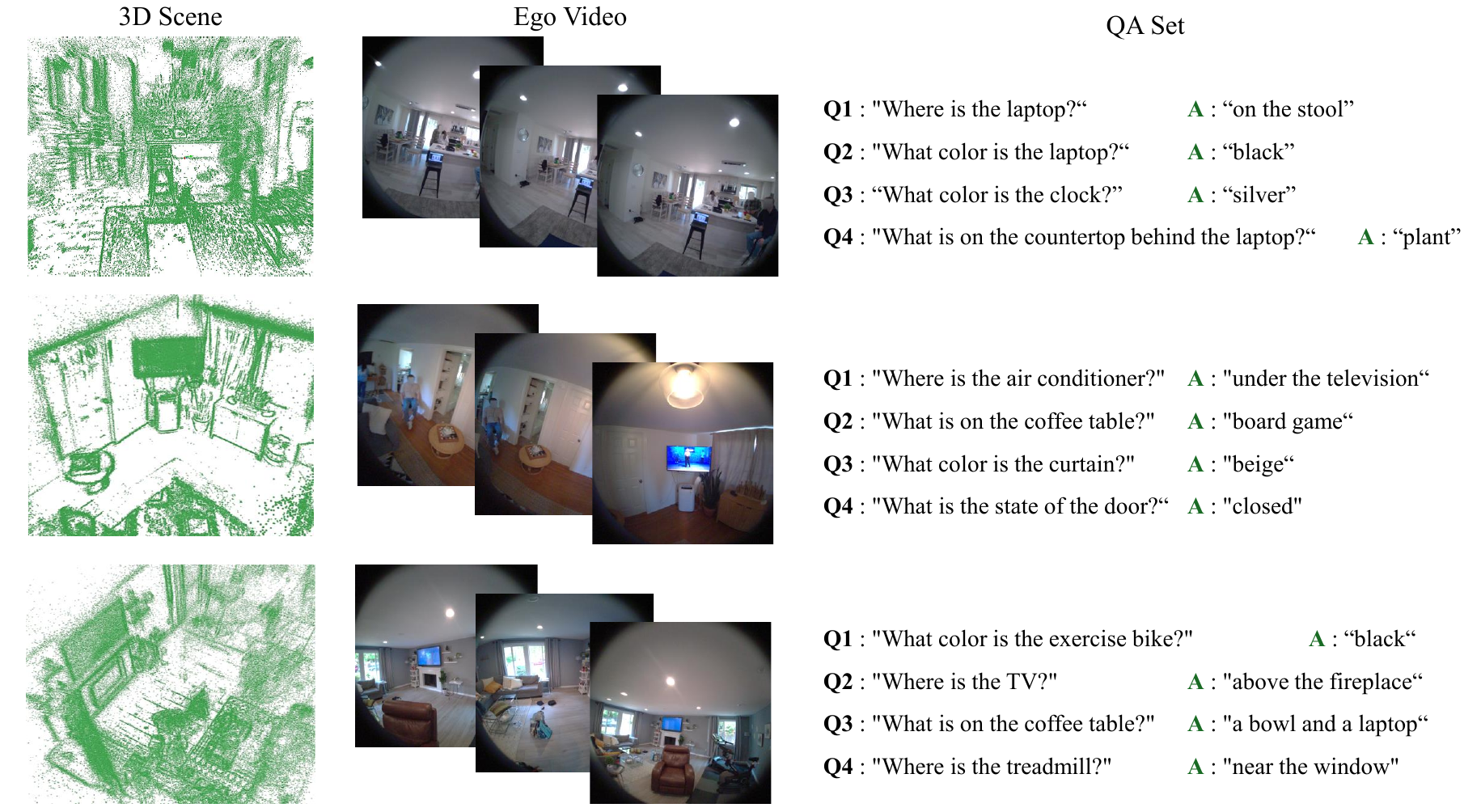} 
    \caption{\textbf{Examples of generated semantic awareness dataset.} 3D scene (left), corresponding egocentric video frames (middle), and example QA items (right). The resulting dataset comprises rich contextual samples that are capable of training the LM to learn the 3D semantics.}
    \label{fig:qa_multi_triplets}
\end{figure*}

We construct a semantic awareness QA dataset using Qwen2.5-VL-7B~\cite{bai2025qwen2}. Given a 3D scene, we sample a short sequence of egocentric frames from each video segment that observes the scene and feed these frames to the model with an instruction prompt as shown below. For every frame, the model produces \(k\) QA items, which we later aggregate over all segments to form a single QA set per scene. The instruction includes (i) a target distribution over QA types (\eg object, color, object nature, place, number, other), (ii) specificity rules that reduce ambiguity by requiring concrete object names, informative modifiers (color/material/shape/size/state), and explicit locator phrases (relation + anchor/reference), and (iii) constraints that exclude humans and human-related attributes. For clarity and consistency, the prompt also provides a few illustrative examples for each QA type. Finally, all QA items generated for the segments associated with a scene are merged and lightly filtered to yield the scene-level semantic QA set. Representative examples of the resulting QA sets are shown in \cref{fig:qa_multi_triplets}. 

\begin{tcolorbox}[colback=white, colframe=black, sharp corners, boxrule=0.5pt, fontupper=\small, breakable,
left=3mm, 
right=3mm, 
]
    \textbf{Semantic Awareness Dataset Generation} \\
    \textbf{Instruction:} \textit{Given the following frames of a 3D scene, generate question–answer (QA) items about the scene. Create {$k$} items in total.}\\
    \textit{Requirements:
    \begin{itemize}
        \item QA type distribution goal: 30\% object, 10\% color, 10\% object nature, 30\% place, 10\% number, 10\% other.
        \item QA specificity: Use concrete object names and, when helpful, modifiers (color/material/shape/size/state) and explicit locator phrases (relation + anchor/reference) to uniquely identify the target.
        \item[]\makebox[\linewidth]{\large$\vdots$}
        \item Exclude humans: If humans appear in the frame, ignore them and focus on non-human objects.
    \end{itemize}}
    \textit{Examples:
    \begin{itemize}
        \item {\textbf{Object}\\
        Q:``What is on $<$Anchor$>$ beside $<$Reference$>$?"\\
        A: ``$<$Target$>$"\\
        Q:``What is on $<$Anchor$>$ to the left of $<$Reference$>$?"\\
        A: ``$<$Target$>$"
        }
        \item {\textbf{Place}\\
        Q:``Where is $<$Target$>$?"\\
        A: ``On $<$Anchor$>$ beside $<$Reference$>$"\\
        Q:``Where is $<$Target$>$ on $<$Anchor$>$?"\\
        A: ``Left of $<$Reference$>$"}
        \item {\textbf{Color}\\
        Q:``What color is $<$Target$>$?"\\
        A: ``$<$Color$>$}
        \item {\textbf{Object nature}\\
        Q:``What material is $<$Target$>$?"\\
        A: ``$<$Material$>$\\
        Q:``What state is $<$Target$>$?"\\
        A: ``$<$State$>$}
        \item {\textbf{Number}\\
        Q:``How many $<$Target$>$ are near $<$Reference$>$?"\\
        A: ``$<$Number$>$}
        \item {\textbf{Other}\\
        Q:``Is there a $<$Target$>$ on $<$Anchor$>$?"\\
        A: ``$<$Yes / No$>$}
    \end{itemize}}
    \textbf{Output:} \textit{$k$ QA items}
\end{tcolorbox}

\begin{figure*}[!t]
    \centering
    \includegraphics[width=1\linewidth]{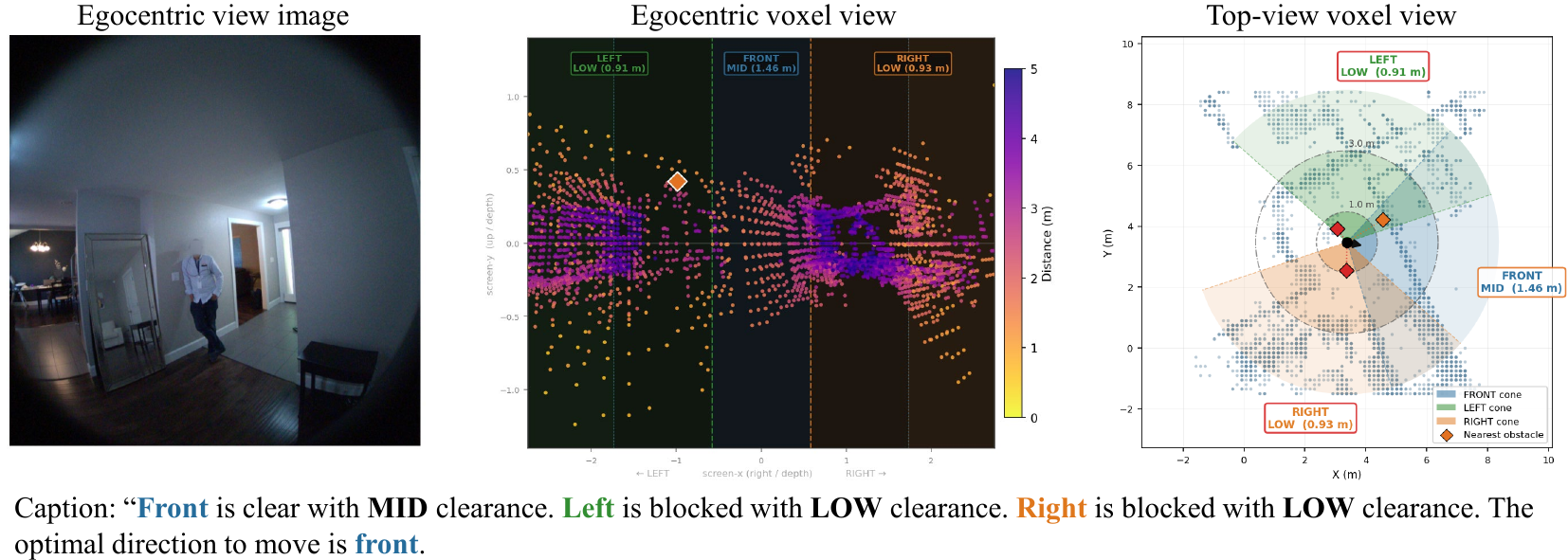} 
    \caption{
    \textbf{Visualization of spatial awareness dataset generation.}
    For each egocentric frame, the surrounding environment is parsed into three directional cone sectors (front, left, right) over the height-filtered 3D voxel reconstruction, and the minimum free-space distance within each cone is discretized into categorical clearance levels (\texttt{LOW}, \texttt{MID}, \texttt{HIGH}) based on the distance to its closest obstacle in each cone sector.
    The resulting spatial caption describes the clearance state and optimal navigation direction, providing structured supervision for the LM to internalize obstacle layouts.}
    \label{fig:qa_obstacle}
\end{figure*}

\subsubsection{Spatial Awareness ($\mathcal{Q}_{\text{spa}}$) Dataset Generation.}
The spatial awareness dataset characterizes the navigatable environment around the human by quantifying directional clearance and collision risk. 
For each observation frame, the dataset provides categorical free-space labels for the front, left, and right directions, corresponding binary collision flags, and a best direction label indicating the least obstructed direction available for locomotion.

Each scene is represented by a set of voxels, a grid origin coordinate, and a uniform voxel resolution of $0.15\mathrm{m}$. 
The head pose from three point tracking provides, for every temporal frame, a 3D translation vector and a $3{\times}3$ rotation matrix.
%
Dataset generation proceeds in two stages. 
First, a proximity filter retains only scene voxels whose vertical offset from the head position falls within $\pm 1.0\mathrm{m}$, restricting the search to body relevant geometry.
Second, for each of the four directions (front, left, right, and back), a cone query selects voxels whose horizontal bearing lies within a cone with a half angle of $60^{\circ}$, and the minimum Euclidean distance to the selected voxels defines the free space distance for that direction. 
Free-space distances are discretized into three categorical levels: \texttt{LOW} ($d < 1.0\mathrm{m}$), \texttt{MID} ($1.0 \leq d < 3.0\mathrm{m}$), and \texttt{HIGH} ($d \geq 3.0\mathrm{m}$). A complementary collision check simulates a $0.5\mathrm{m}$ step in each direction and tests whether any voxel falls within a cylindrical body proxy of radius $0.3\mathrm{m}$ and half-height $1.0\mathrm{m}$ centered at the projected position. The direction with the greatest free space distance is designated as the best direction label (\textsc{Best\_Dir}). The resulting caption summarizes this information in a single sentence, as shown in Fig.~\ref{fig:qa_obstacle}. 
Full algorithmic details are provided in Algorithm~\ref{algo:obstacle}.
  \begin{algorithm}[t]
  \caption{Spatial Awareness Dataset Generation}
  \label{alg:obstacle_parse}
  \begin{algorithmic}[1]
  \Require Scene point cloud $\mathcal{S} \subset \mathbb{R}^3$,
           head position $\mathbf{p} \in \mathbb{R}^3$,
           rotation matrix $\mathbf{R} \in \mathrm{SO}(3)$, and horizontal projection $\Pi$
  \Ensure  Labels $\{$\texttt{FREE\_FRONT}, \texttt{FREE\_LEFT}, \texttt{FREE\_RIGHT},
           \texttt{COLLIDE\_FRONT}, \texttt{COLLIDE\_LEFT}, \texttt{COLLIDE\_RIGHT},  \texttt{BEST\_DIR}$\}$

  \State $\mathbf{f} \leftarrow \Pi(\mathbf{R}[:,1])$,\quad
         $\mathbf{r} \leftarrow \Pi(\mathbf{R}[:,0])$
         \Comment{Horizontal projections of forward/right axes}
  \State $\mathcal{D} \leftarrow \{\mathbf{f},\,-\mathbf{r},\,\mathbf{r},\,-\mathbf{f}\}$
         \Comment{FRONT, LEFT, RIGHT, BACK unit vectors}
  \State $\mathcal{S}' \leftarrow \{\mathbf{s} \in \mathcal{S} :
         |s_z - p_z| < \delta_h\}$
         \Comment{Height filter: $\delta_h = 1.0\,\mathrm{m}$}
  \For{each direction $\mathbf{d} \in \mathcal{D}$}
      \State $\mathbf{r}_i \leftarrow \mathbf{s}_i - \mathbf{p}$,\quad
             $d_i \leftarrow \|\mathbf{r}_i\|$ \quad $\forall\,\mathbf{s}_i \in \mathcal{S}'$
      \State $\hat{\mathbf{r}}^h_i \leftarrow \Pi(\mathbf{r}_i)\,/\,\|\Pi(\mathbf{r}_i)\|$
             \Comment{Horizontal unit direction to voxel}
      \State $\mathcal{C} \leftarrow \{i : \hat{\mathbf{r}}^h_i \cdot \mathbf{d} > \cos\theta_c,\;
             0.05 < d_i < d_{\max}\}$
             \Comment{Cone filter: $\theta_c = 60^\circ$}
      \State $\mathrm{free}(\mathbf{d}) \leftarrow
             \min_{i \in \mathcal{C}} d_i$
             \quad (or $d_{\max}$ if $\mathcal{C} = \varnothing$)
      \State $\mathrm{category}(\mathbf{d}) \leftarrow
             \begin{cases}
               \texttt{LOW}  & \text{if } \mathrm{free}(\mathbf{d}) < 1.0 \\
               \texttt{MID}  & \text{if } 1.0 \le \mathrm{free}(\mathbf{d}) < 3.0 \\
               \texttt{HIGH} & \text{otherwise}
             \end{cases}$
      \State $\mathbf{p}' \leftarrow \mathbf{p} + \Delta s\,\mathbf{d}$
             \Comment{Simulated step: $\Delta s = 0.5\,\mathrm{m}$}
      \State $\mathrm{collide}(\mathbf{d}) \leftarrow
             \exists\,\mathbf{s} \in \mathcal{S} :
             \|\Pi(\mathbf{s} - \mathbf{p}')\| < r_b
             \;\wedge\; |s_z - p'_z| < \delta_h$
             \Comment{$r_b = 0.3\,\mathrm{m}$}
  \EndFor
  \State $\textsc{Best\_Dir} \leftarrow
         \arg\max_{\mathbf{d} \in \mathcal{D}}\;\mathrm{free}(\mathbf{d})$
  \State \Return categorical labels and collision flags for FRONT/LEFT/RIGHT, \textsc{Best\_Dir}
  \end{algorithmic}
  \label{algo:obstacle}
  \end{algorithm}

\subsection{Data Split}
We conduct training and evaluation on the Nymeria dataset~\cite{ma2024nymeria}, using the provided human motion, three-point trackings, motion descriptions, 3D scene point clouds, and egocentric video frames. As Nymeria \emph{does not offer an official split}, we construct splits based on scenes containing narration annotations. 
We stratify scenes by dominant motion pattern (largely static vs. dynamic human motion) to maintain a similar static/dynamic ratio in both splits, based on Nymeria's scene activity tags (action scripts); for example, static activities include \textit{simon says}, \textit{laundry}, and \textit{cooking}, while dynamic activities include \textit{housekeeping}, \textit{welcome to my place}, and \textit{where is X}.
We set a ratio of 4:1 between training and validation scenes. 
The exact split specification will be released alongside our code.

\subsection{Baseline Models}
\noindent\textbf{EgoLM.}
We adapt EgoLM~\cite{hong2025egolm}, originally designed for egocentric motion tracking and narration, to motion forecasting. Because EgoLM does not directly support forecasting instructions, we instead use its motion-to-motion (M2M) pretraining, in which the language model backbone is trained to autoregressively generate motion sequences.

We consider two EgoLM baselines that differ in how the past motion condition is obtained. \textbf{EgoLM (GT motion)} uses the M2M pretrained checkpoint and conditions forecasting on the ground truth past motion. \textbf{EgoLM (Inst. tuning)} first uses the instruction tuned EgoLM checkpoint to reconstruct past motion from sparse three point tracking, and then feeds the reconstructed past motion to the same M2M pretrained checkpoint to forecast the future.

For both variants of the inference, directly prompting the model with the full past sequence often leads to early termination and truncated outputs. To mitigate this implicit length bias and generate the full prediction horizon, we use a recursive two-step generation strategy. We first condition on the latter half of the past motion to generate the first half of the future motion, and then condition on the generated segment to produce the remaining second half of the future.


    


\noindent\textbf{FIction.}
FIction~\cite{ashutosh2025fiction} originally predicts future 3D human pose sequence from past 3D human pose sequence, given past sequence of egocentric video and 3D scene represented as voxel. 
We modify this framework to our setting, namely: (i) predicting future 3D human pose sequence from past three-point tracking data and (ii) tracking past 3D pose sequence from three-point tracking data, where egocentric video and 3D scene representations are provided.
Therefore, we modify the encoder to encode three-point tracking data instead of the past 3D human pose sequence via an MLP network. 
Also, we additionally add a past 3D pose sequence tracking head decoder that still samples based on CVAE, as in the original future pose prediction head.

\noindent\textbf{UniEgoMotion.}
UniEgoMotion~\cite{patel2025uniegomotion} is a unified conditional motion diffusion model that reconstructs, forecasts, and generates scene-aware 3D human body motion from egocentric images alone.
We re-implement UniEgoMotion to take 3D scene and three-point tracking data as input in a similar manner as FIction and our Ego3DLM.
Also, past egocentric video is given as input instead of single image.

\noindent\textbf{LLM Narration.}
As no works exist that predict future motion in text modality as in our predictive setting, we compare with the zero-shot capability of Qwen2.5-VL-7B~\cite{bai2025qwen2}. We provide uniformly sampled 10 egocentric video frames from the past sequence and prompt the model to predict the future motion description with the following prompt.

\begin{tcolorbox}[colback=white, colframe=black, sharp corners, boxrule=0.5pt, fontupper=\small, breakable,
left=3mm, 
right=3mm, 
]
    \textbf{Instruction:} \textit{You are given RGB frames from sequence $<$source\_seq$>$ of scene $<$scene\_name$>$. \\
    Predict what is most likely to happen in the immediate next sequence $<$target\_seq$>$. \\
    Focus on motion and scene dynamics, not static descriptions. \\}

    \textit{Study the following ground-truth style examples and mimic their tone, tense, and specificity: $<$examples\_block$>$ \\}

    \textit{Style constraints:
    \begin{itemize}
        \item Always describe the actor as``C" (\eg``C is standing in the hallway...").
        \item Use present/progressive tense to mirror GT descriptions (avoid ``will/going to").
        \item Include concrete verbs and referenced objects/surfaces whenever visible.
        \item 1-3 sentences maximum; no bullet lists.
    \end{itemize}}
\end{tcolorbox}

For the $<$\textit{examples\_block}$>$, we provide the following GT sequences to provide context for the LLM to generate responses with a similar style of text as the GT:
\textit{
\begin{itemize}
    \item ``C is standing in the hallway, then leans to her right to reach for the door before opening it with her right hand."
    \item ``C walks into the kitchen, rests her left hand on the counter, and bends to peer into the refrigerator."
    \item ``C steps toward the sofa, lowers herself into a seated position, and keeps talking to the person next to her."
\end{itemize}
}

\subsection{Evaluation Metric}
All metrics calculated in the embedding space (FID, Diversity, $d_{gp}, d_{pg}, d_{pp}$) are first obtained the motion and text embedding with respective encoders trained as in~\cite{guo2022generating} on our training split of the Nymeria dataset.

\noindent\textbf{Common Motion Metric.}
\noindent As local motion plausibility plays a critical role in both motion forecasting and tracking, we employ APE as a shared metric to effectively measure the local positional accuracy of the generated motions.
\begin{itemize}
    \item \noindent\textbf{APE (Aligned mean per-joint Position Error):}
APE measures local pose accuracy as the mean $\ell_2$ distance between predicted and ground-truth joint positions after aligning the head position of both predicted and ground-truth skeletons, evaluating articulated motion independent of global translation.
\end{itemize}

\noindent\textbf{Motion Forecasting Metric.}
\noindent For the future motion prediction, we generate three modes and select the best one based on the lowest JPE, since it comprehensively captures both global translation and local joint articulation errors. To evaluate joint positional accuracy, we calculate APE, JPE, and ADE, alongside FDE to assess the precision of the final displacement. For both ADE and FDE, we also provide short-term evaluations restricted to the first 2 seconds. Additionally, we compute FID and Diversity to evaluate the statistical realism and variation of the generated motion distributions.
\begin{itemize}
    \item \noindent\textbf{JPE (Joint Position Error):}
JPE measures global pose accuracy as the mean $\ell_2$ distance between predicted and ground-truth joint positions in the world frame, penalizing errors in both body placement and joint configuration.
    \item \noindent\textbf{ADE (Average Displacement Error of head position):}
ADE is the average Euclidean distance between predicted and ground-truth head positions over all forecast timesteps, analogous to the standard ADE metric in trajectory prediction. $\text{ADE}_{2s}$ refers to ADE of the first 2 seconds.
    \item \noindent\textbf{FDE (Final Displacement Error of head position):}
FDE is the Euclidean distance between predicted and ground-truth head positions at the final forecast timestep, emphasizing long-horizon accuracy.
$\text{FDE}_{2s}$ refers to FDE at 2 seconds.
    \item \noindent\textbf{FID (Fréchet Inception Distance on pose sequences):}
FID is computed between multivariate Gaussian fits of ground-truth and generated pose-sequence embeddings (random 300 validation samples), with lower values indicating a closer match between real and generated motion distributions.
    \item \noindent\textbf{Diversity (pose-sequence embedding diversity):}
Diversity is the mean $\ell_2$ distance between randomly sampled pairs of pose-sequence embeddings (up to 300 pairs), where higher values indicate a richer, more varied set of generated motions.\\
\end{itemize}

\noindent\textbf{Motion Tracking Metric.}
\noindent For motion tracking evaluation, we evaluate positional accuracy using the overall APE, as well as the regional Upper and Lower body APE. Furthermore, we assess rotational accuracy by calculating the root joint angular error and the mean rotation error across all articulated joints.
\begin{itemize}
    \item \noindent\textbf{Upper and Lower (Upper \& Lower body APE):}
Upper and Lower measure local pose accuracy by calculating the mean APE for specific body parts. Upper is the mean APE of 14 joints, including the root joint and those located above it, while Lower is the mean APE of 10 joints, including the root joint and those located below it.
    \item \noindent\textbf{Root (Root Angle):}
Root angle is the mean geodesic angular error between the root orientations of the aligned predicted past poses and the ground-truth past poses, reflecting how accurately the pelvis/torso facing direction is recovered.
    \item \noindent\textbf{J.A. (Joint Angle):}
Joint angle is the mean rotation error over all articulated joints: each child-parent bone is converted to a local 6D rotation and compared with the corresponding ground-truth joint frame, measuring how well limb and joint orientations match independently of the root pose.
\end{itemize}

\noindent\textbf{Text Metric.}
\noindent In motion description evaluation, we use widely adopted metrics for generated text quality, including Bleu, $\text{Rouge}_L$, SBert, and R@N.
\begin{itemize}
    \item \noindent\textbf{Bleu and $\text{Rouge}_L$:}
Bleu~\cite{papineni2002bleu} measures $n$-gram overlap between predicted and ground-truth texts to quantify lexical precision, scaled from 0 to 1. $\text{Rouge}_L$~\cite{lin2004rouge} evaluates the Longest Common Subsequence (LCS) between predicted texts and the ground truth, effectively capturing similarities in word order and overall sentence structure. Please refer to the corresponding papers for detailed mathematical formulations.
    \item \noindent\textbf{SBert (SBert similarity between sentences):}
SBert~\cite{reimers2019sentence} measures the semantic alignment between predicted and ground-truth texts by computing the mean cosine similarity of their sentence embeddings encoded via a pre-trained SBERT model.
    \item \noindent\textbf{R@N (top-N R-precision accuracy):}
R@N evaluates the Euclidean distances between predicted text embeddings and ground-truth motion embeddings within a fixed-size subset of size 300. It represents the proportion of instances where the true corresponding ground-truth motion embedding is ranked within the top $N$ closest matches to the predicted text embedding, with higher values indicating stronger semantic alignment between the text and motion modalities.\\
\end{itemize}

\noindent\textbf{Motion-Text Alignment Metric.}
To evaluate the semantic consistency and cross-modal alignment between textual descriptions and generated motions, we utilize narration-motion matching distances.
\begin{itemize}
    \item \noindent\textbf{$\boldsymbol{d}_{gp}$, $\boldsymbol{d}_{pg}$, $\boldsymbol{d}_{pp}$ (Narration-motion matching distance):}
Narration-motion matching distance measures the cross-modal alignment as the mean $\ell_2$ distance between narration and motion embeddings. It evaluates the semantic correspondence across different pairings: ground-truth narration versus predicted motion ($d_{gp}$), predicted narration versus ground-truth motion ($d_{pg}$), and the self-consistency of jointly predicted narration-motion pairs ($d_{pp}$).
\end{itemize}

\section{Additional Experimental Results}
\label{sec:exp_supp}
In this section, we provide additional analyses of our design choices and the effects of 3D scene conditioning. 
We compare baselines with and without 3D scene inputs, and ablate joint decoding across motion and text as well as across past and future tasks, highlighting consistent gains from unified generation. 
We further evaluate long-term forecasting and multi-hypothesis future prediction to assess effectiveness across extended horizons and generation diversity. 
Ego3DLM$^\dagger$ denotes our model trained up to instruction tuning only, without the GRPO reinforcement learning stage, to directly isolate and clarify the impact of our core model design.

\begin{table}[t]
    \centering
    \caption{Ablation on the effect of 3D scene conditioning across all methods on motion prediction and tracking. While all methods benefit from 3D scene input, Ego3DLM$^{\dagger}$ achieves the largest and most consistent gains.}
    \renewcommand{\arraystretch}{1.3}
    \setlength{\tabcolsep}{4pt}
    \resizebox{\linewidth}{!}{%
        \begin{tabular}{lc cccccc ccccc}
        \toprule
        \multirow{2}{*}{Method} & \multirow{2}{*}{3D Scene} & \multicolumn{6}{c}{Motion Prediction (3 modes)} & \multicolumn{5}{c}{Motion Tracking} \\
        \cmidrule(lr){3-8} \cmidrule(lr){9-13}
         & & APE $\downarrow$ & JPE $\downarrow$ & ADE $\downarrow$ & FDE $\downarrow$ & FID $\downarrow$ & Diversity $\uparrow$ & APE $\downarrow$ & Upper $\downarrow$ & Lower $\downarrow$ & J.A. $\downarrow$ & Root $\downarrow$ \\
        \midrule
        \multirow{2}{*}{Fiction} & & 209.5 & 587.5 & 582.5 & 949.6 & 0.3691 & 0.7381 & 187.5 & 117.6 & 285.7 & 32.94 & 35.11 \\
         & \checkmark & 206.2 & 564.7 & 558.3 & 904.3 & 0.3275 & 0.7432 & 181.1 & 114.0 & 282.3 & 31.28 & 34.61 \\
        \midrule
        \multirow{2}{*}{\makecell[l]{EgoLM (Inst. tune)}} & & 184.9 & 579.4 & 552.6 & 963.6 & 0.2137 & 0.6142 & 161.9 & 95.6 & 265.6 & 33.63 & 24.01 \\
         & \checkmark & 174.2 & 493.1 & 471.2 & 925.2 & 0.2184 & 0.5176 & 148.5 & 83.7 & 232.6 & 34.14 & 30.64 \\
        \midrule
        \multirow{2}{*}{UniEgoMotion} & & 157.5 & 439.5 & 430.4 & 740.6 & 0.1494 & 0.9171 & 156.2 & 82.7 & 248.5 & 28.52 & 23.96 \\
         & \checkmark & 151.5 & 424.3 & 409.7 & 720.5 & 0.1530 & 0.9022 & 152.2 & 79.4 & 233.3 & 26.48 & 22.71 \\
        \midrule
        \multirow{2}{*}{$\text{Ego3DLM}^{\dagger}$} & & 156.2 & 388.5 & 363.9 & 678.4 & 0.0426 & 0.9352 & 105.8 & 57.5 & 168.8 & 24.03 & 20.92 \\
         & \checkmark & \textbf{148.5} & \textbf{368.3} & \textbf{346.3} & \textbf{645.4} & \textbf{0.0190} & \textbf{0.9358} & \textbf{98.3} & \textbf{53.4} & \textbf{155.3} & \textbf{22.52} & \textbf{19.62} \\
        \bottomrule
        \end{tabular}%
    }
    \label{supp:modality_wise}
    \vspace{-1.0em}
\end{table}
  
\noindent\textbf{Additional Baselines with/without 3D Scene Conditioning.}
Table~\ref{supp:modality_wise} evaluates all methods with and without 3D scene input on motion prediction and tracking. 
While all baselines show modest JPE improvements from 3D scene conditioning, with Fiction reducing JPE by $3.9\%$ and UniEgoMotion by $3.5\%$, these gains are limited by the absence of dedicated scene understanding, as scene features are injected without prior spatial or semantic grounding. In contrast, Ego3DLM$^\dagger$ achieves a larger JPE reduction of $5.2\%$ with scene conditioning and reaches a JPE of $368.3$, a $13.2\%$ improvement over the strongest baseline, UniEgoMotion with 3D scene ($424.3$). We attribute this gain to our spatial-semantic pretraining, which explicitly trains the model to reason about object-level geometry and semantics before motion prediction is introduced, enabling the model to translate 3D scene geometry into physically informed and spatially coherent body motion predictions.

\noindent\textbf{Separate Motion/Text Decoding.}
Table~\ref{supp:motion_desc_updated} shows that jointly decoding motion and text in a single autoregressive sequence consistently improves accuracy and narration quality over separate motion and text decoding. Compared to the separated variant, joint decoding reduces motion prediction errors by 0.67\% in APE, 0.51\% in JPE, and 2.26\% in FDE.
It also improves future motion description, increasing Bleu-4 by 3.19\% and $\text{Rouge}_{L}$ by 2.42\%, with retrieval gains of 1.96\% in R@3 and 4.59\% in R@1. Critically, narration motion alignment strengthens, with $d_{pg}$ decreasing by 1.35\% and $d_{pp}$ dropping by 9.25\%, supporting that joint generation yields more mutually grounded pose narration pairs.

\noindent\textbf{Separate Past/Future Decoding.}
Table~\ref{supp:motion_tracking} shows that jointly decoding past and future tasks improves both forecasting and tracking compared to separate decoding. For motion prediction, joint decoding reduces JPE by 7.18\%, with consistent improvements across the other forecasting metrics. For motion tracking, it yields larger gains, reducing APE by 28.04\%, with consistent gains across the other tracking metrics. These results suggest that training the model to produce past and future motion within a unified generation process encourages shared temporal representations that benefit both tracking fidelity and forecasting performance. This joint supervision appears to strengthen the model’s understanding of motion dynamics, improving the fidelity of generated motion and enabling forecasting that more effectively leverages past observations to produce feasible future predictions.

\newcommand{\vc}[1]{\raisebox{0.5\normalbaselineskip}[0pt][0pt]{#1}}

\begin{table}[t]
    \centering
    \caption{Ablation study on joint vs.\ separate motion and text decoding. Joint decoding consistently improves motion prediction accuracy and future motion description quality, while also strengthening pose-narration alignment.}
    \renewcommand{\arraystretch}{1.3}
    \setlength{\tabcolsep}{4pt}
    \resizebox{\linewidth}{!}{%
        \begin{tabular}{lcccccc ccccc c}
        \toprule
        \multirow{2}{*}{Method} & \multicolumn{6}{c}{Motion Prediction (3 modes)} & \multicolumn{5}{c}{Future motion description} & \multicolumn{1}{c}{x-y align} \\
        \cmidrule(lr){2-7} \cmidrule(lr){8-12} \cmidrule(lr){13-13}
         & APE $\downarrow$ & JPE $\downarrow$ & ADE $\downarrow$ & FDE $\downarrow$ & FID $\downarrow$ & Diversity $\uparrow$ & Bleu-4 $\uparrow$ & Rogue$_{L}$ $\uparrow$ & \makecell[c]{R@3 $\uparrow$} & \makecell[c]{R@1 $\uparrow$} & \makecell[c]{$d_{pg}$ $\downarrow$} & $d_{pp}$ $\downarrow$ \\
        \midrule
        \shortstack[l]{$\text{Ego3DLM}^{\dagger}$ \\(sep. motion/text)} & \vc{149.5} & \vc{370.2} & \vc{348.7} & \vc{660.3} & \vc{\textbf{0.0116}} & \vc{\textbf{1.0260}} & \vc{0.1004} & \vc{0.2971} & \vc{0.2654} & \vc{0.1025} & \vc{6.5723} & \vc{6.5201} \\
        $\text{Ego3DLM}^{\dagger}$    & \textbf{148.5} & \textbf{368.3} & \textbf{346.3} & \textbf{645.4} & 0.0190 & 0.9358 & \textbf{0.1036} & \textbf{0.3043} & \textbf{0.2706} & \textbf{0.1072} & \textbf{6.4837} & \textbf{5.917} \\
        \bottomrule
        \end{tabular}%
    }
    \label{supp:motion_desc_updated}
\end{table}

\begin{table}[t]
    \centering
    \caption{Ablation study on joint vs.\ separate past and future task decoding. Joint decoding consistently improves both motion prediction and tracking.
    }
    \renewcommand{\arraystretch}{1.3}
    \setlength{\tabcolsep}{4pt}
    \resizebox{\linewidth}{!}{%
        \begin{tabular}{lcccccc ccccc}
        \toprule
        \multirow{2}{*}{Method} & \multicolumn{6}{c}{Motion prediction (3 modes)} & \multicolumn{5}{c}{Motion tracking} \\
        \cmidrule(lr){2-7} \cmidrule(lr){8-12}
         & APE $\downarrow$ & JPE $\downarrow$ & ADE $\downarrow$ & FDE $\downarrow$ & FID $\downarrow$ & Diversity $\uparrow$ & APE $\downarrow$ & Upper $\downarrow$ & Lower $\downarrow$ & J.A. $\downarrow$ & Root $\downarrow$ \\
        \midrule
        \shortstack[l]{$\text{Ego3DLM}^{\dagger}$ \\(sep. past/future)} & \vc{159.9} & \vc{396.8} & \vc{370.5} & \vc{705.0} & \vc{0.0335} & \vc{\textbf{0.9876}} & \vc{136.6} & \vc{77.5} & \vc{213.2} & \vc{31.08} & \vc{29.00} \\
        $\text{Ego3DLM}^{\dagger}$    & \textbf{148.5} & \textbf{368.3} & \textbf{346.3} & \textbf{645.4} & \textbf{0.0190} & 0.9358 & \textbf{98.3} & \textbf{53.4} & \textbf{155.3} & \textbf{22.5} & \textbf{19.62} \\
        \bottomrule
        \end{tabular}%
    }
    \label{supp:motion_tracking}
\end{table}

\noindent\textbf{Long-Term Motion Forecasting.}
\label{sec:long_term_supp}
Table~\ref{supp:longterm} evaluates all methods on long-term motion prediction, where the increased temporal horizon amplifies the compounding effect of errors and scene-dependency of future motion.
Ego3DLM$^\dagger$ consistently outperforms all baselines across every metric, reducing JPE by $18.7\%$ and ADE by $19.2\%$ over the strongest baseline UniEgoMotion, demonstrating that our scene-aware training is particularly beneficial as the prediction horizon grows.
Notably, EgoLM variants perform poorly even with ground-truth past motion as input, suggesting that neither language grounding nor GT conditioning alone is sufficient for accurate long-term forecasting without explicit 3D scene understanding.

\begin{table}[t]
    \centering
    \caption{Long-term motion prediction results across all methods. Ego3DLM$^\dagger$ achieves state-of-the-art performance across all metrics, with the performance gap over baselines widening compared to the standard horizon.
    }
    \renewcommand{\arraystretch}{1.3}
    \setlength{\tabcolsep}{4pt}
    \resizebox{0.7\linewidth}{!}{%
        \begin{tabular}{lcccccc}
        \toprule
        \multirow{2}{*}{Method} & \multicolumn{6}{c}{Prediction (3 modes)} \\
        \cline{2-7}
         & APE $\downarrow$ & JPE $\downarrow$ & ADE $\downarrow$ & FDE $\downarrow$ & FID & Diversity \\
        \midrule
        Fiction & 256.1 & 831.1 & 825.7 & 1,083.9 & 0.0896 & 0.3815 \\
        \makecell[l]{EgoLM (GT)} & 242.2 & 917.1 & 884.2 & 1,469.3 & 0.1086 & 0.2275 \\
        \makecell[l]{EgoLM (Inst.\ tune.)} & 238.6 & 922.3 & 887.4 & 1,472.9 & 0.1095 & 0.2231 \\
        UniEgoMotion & 201.7 & 784.7 & 765.1 & 1,183.4 & 0.0858 & 0.3981 \\
        $\text{Ego3DLM}^{\dagger}$ & \textbf{172.2} & \textbf{638.5} & \textbf{618.3} & \textbf{1,079.8} & \textbf{0.0325} & \textbf{0.4141} \\
        \bottomrule
        \end{tabular}%
    }
    \label{supp:longterm}
    \vspace{-1.0em}
\end{table}

\begin{figure*}[!t]
    \centering
    \includegraphics[width=1\linewidth]{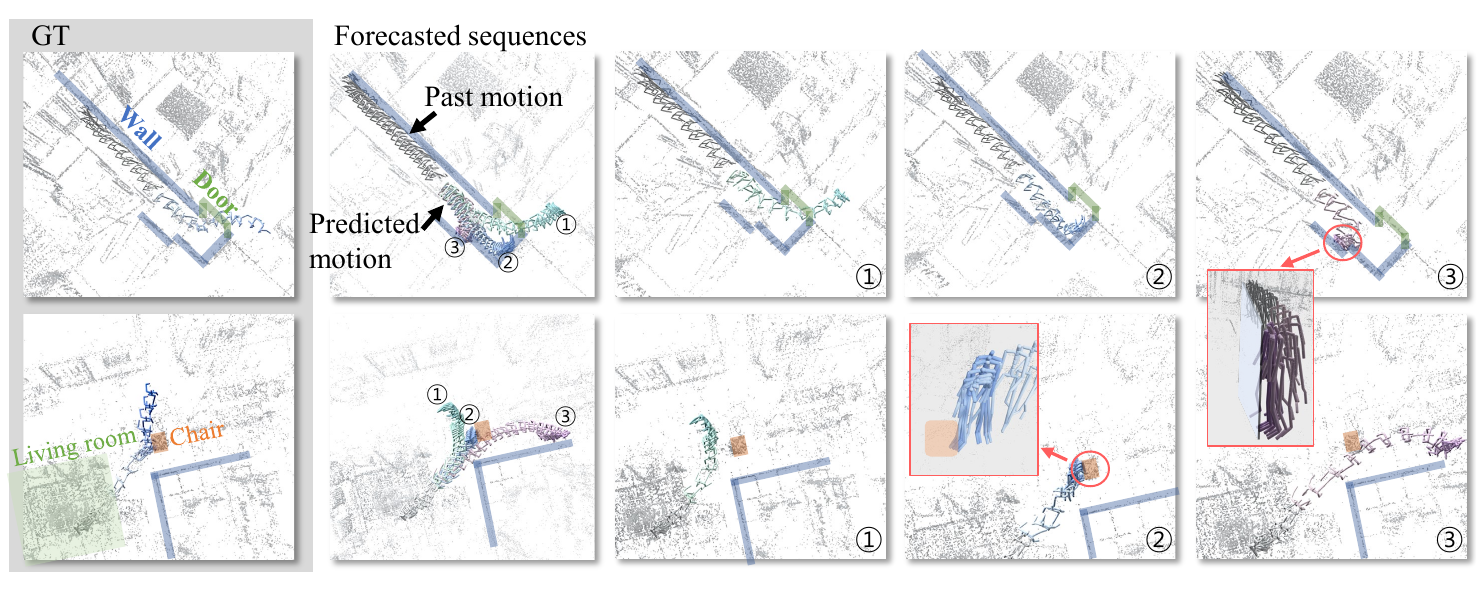} 
    \caption{\textbf{Examples of multi-hypothesis predictions.} GT motion sequence (left), forecasted motion sequences (middle \& right). All our predictions demonstrate feasible motion based on the 3D scene.}
    \label{fig:qual_mm}
\end{figure*}

\noindent\textbf{Multi-Hypothesis Motion Forecasting.}
\label{sec:multi_modal_supp}
\Cref{fig:qual_mm} visualizes the multi-hypothesis predictions generated by our model.
Human motion is inherently non-deterministic, as a person may exhibit different plausible behaviors given the same past motion depending on their underlying intent.
To address this property, we sample three distinct future predictions from the same model and examine the 3D human motion fidelity of each.
Across all samples, the predicted trajectories remain plausible within the 3D scene, maneuvering through the indoor environment while avoiding or interacting with objects.
These results indicate that our model effectively captures the non-deterministic nature of human motion intentions, enabled by its ability to leverage the rich semantic cues provided by the 3D scene information.

\section{Additional Qualitative Results}
\label{sec:qual_supp}
\begin{figure*}[!t]
    \centering
    \includegraphics[width=0.9\linewidth]{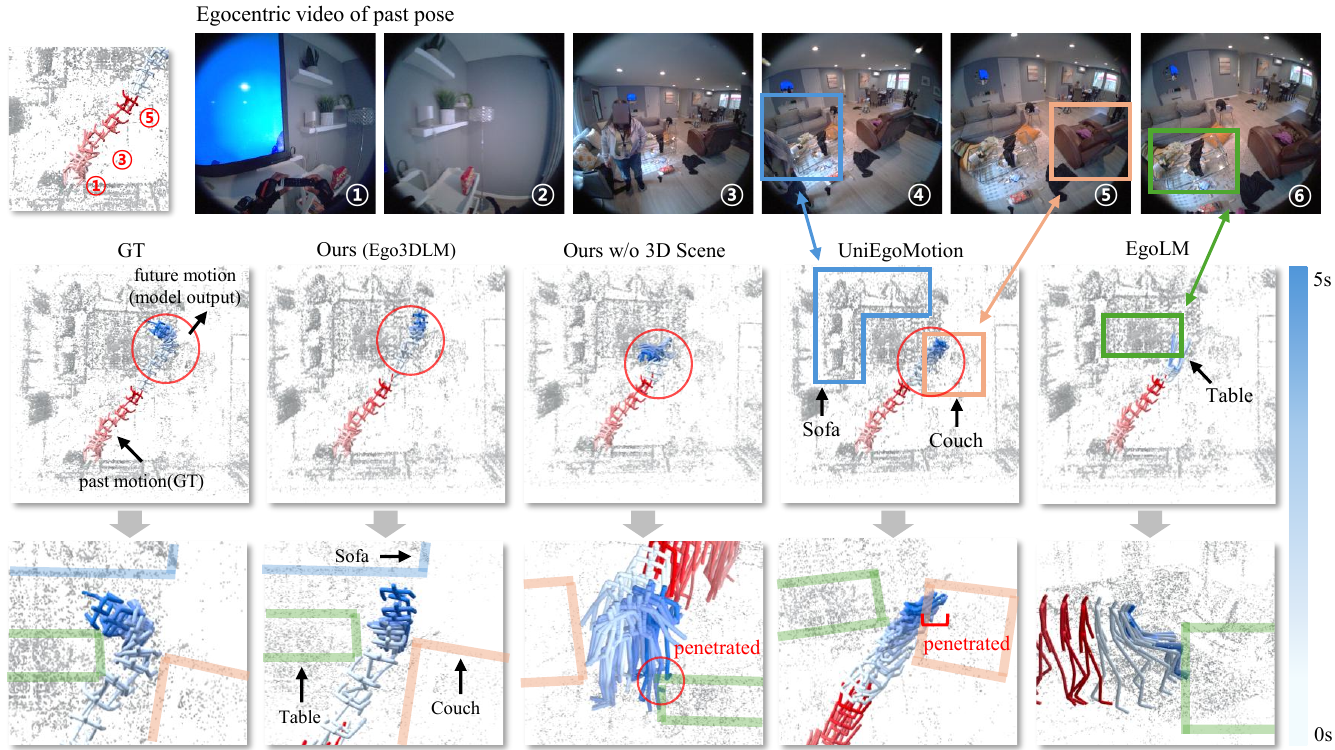} 
    \par \vspace{0.5cm} 
    
    \includegraphics[width=0.89\linewidth]{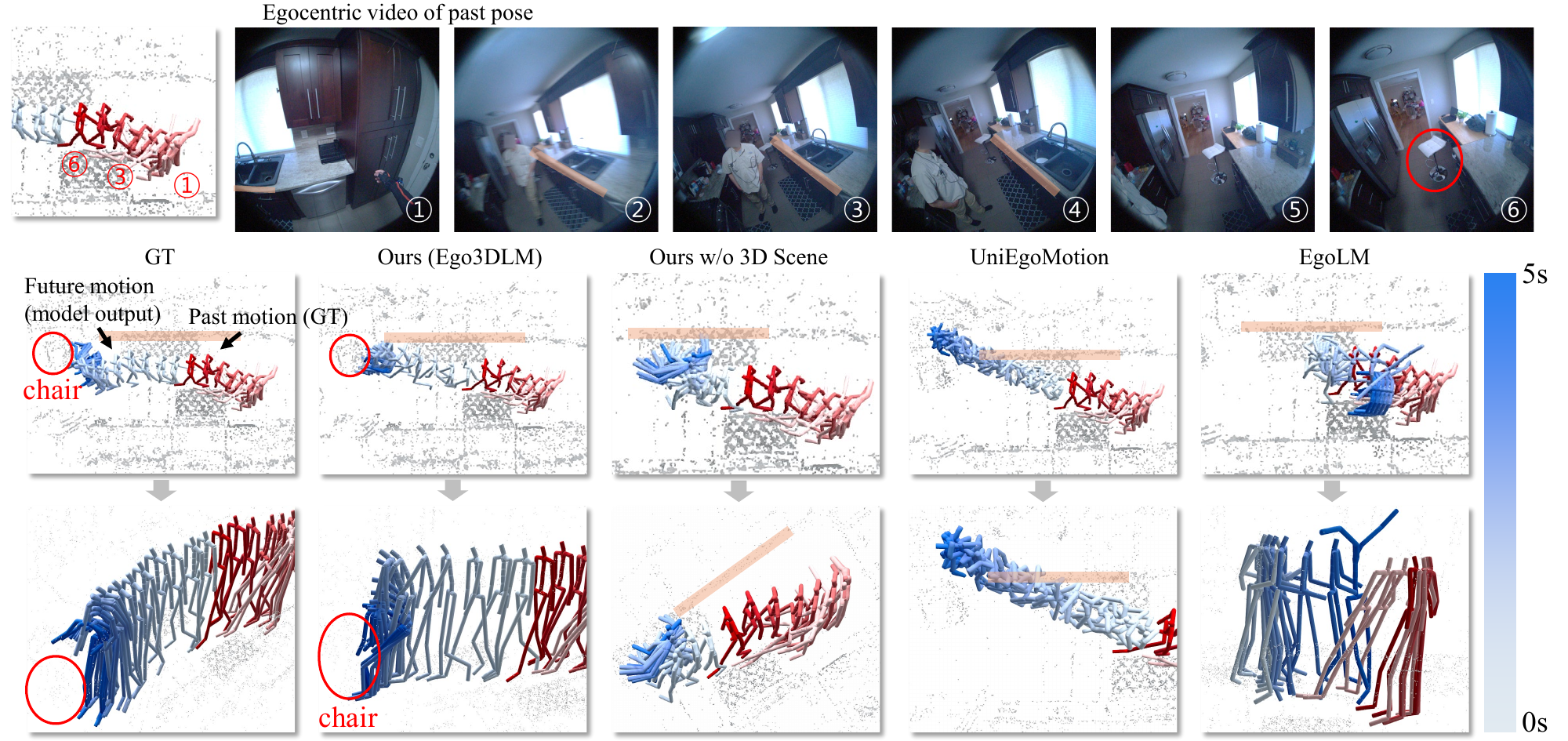} 
    
    \caption{
        \textbf{Qualitative results of motion prediction and future motion description}. 
        We visualize predicted motions from white to blue with temporal progression. 
        Ego3DLM generates plausible trajectories that respect spatial boundaries, whereas baselines and the ablation variant frequently collide with scene geometry.
    }
    \label{fig:qual_sup_pred_1_2}
\end{figure*}

\begin{figure*}[!t]
    \centering
    \includegraphics[width=0.9\linewidth]{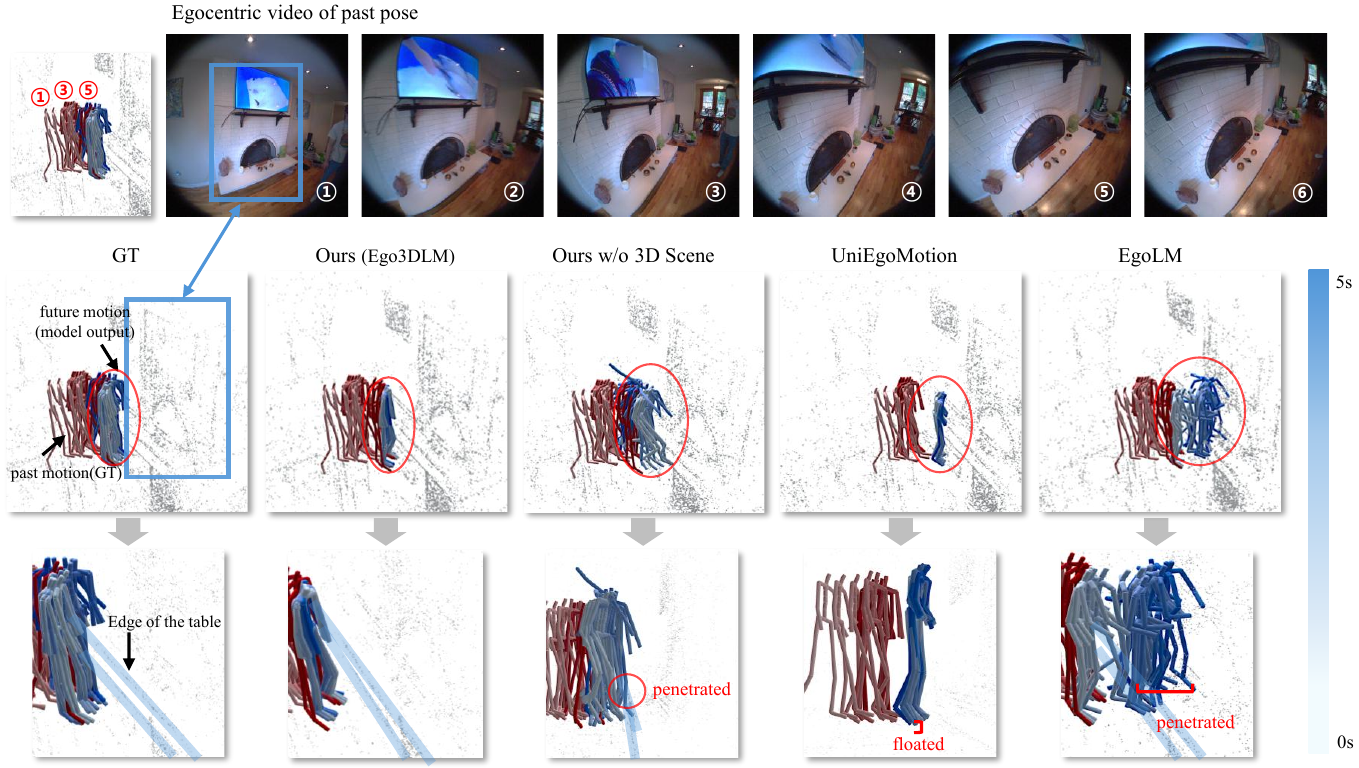} 
    \par \vspace{0.5cm}
    
    \includegraphics[width=0.89\linewidth]{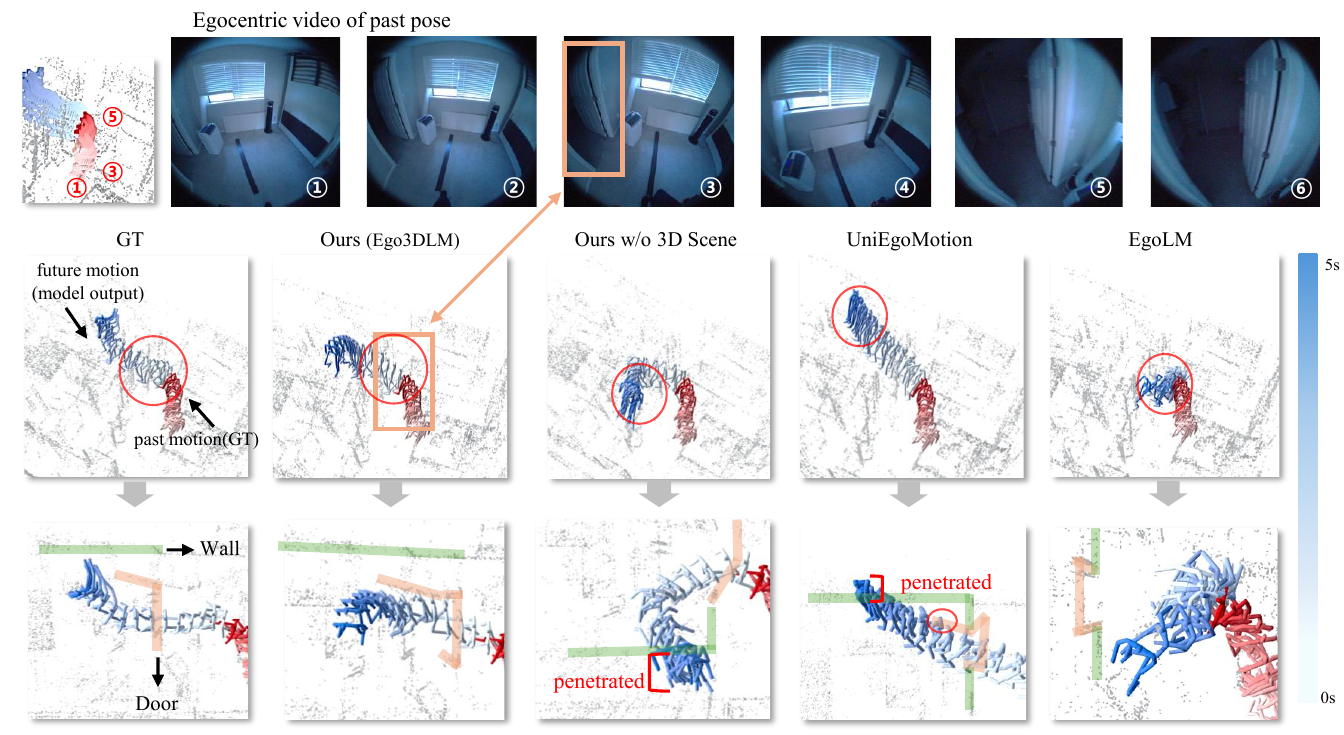} 
    
    \caption{
        \textbf{Qualitative results of motion prediction and future motion description}. 
        We visualize predicted motions from white to blue with temporal progression. 
        Ego3DLM produces plausible trajectories that adhere to scene boundaries, while the baselines and ablated variant frequently intersect or collide with the surrounding geometry.
    }
    \label{fig:qual_sup_pred_3_4}
\end{figure*}

We present an additional qualitative comparison of our method against the ablation variant (Ours w/o 3D Scene) and baselines across diverse scenarios in \cref{fig:qual_sup_pred_1_2} and \cref{fig:qual_sup_pred_3_4}.
The forecasted motion sequences from $t=0$s to $t=5$s are color-coded to indicate temporal progression, transitioning from white to blue, with the past motion shown in red.
Ego3DLM demonstrates a robust understanding of environmental constraints.
By explicitly conditioning on 3D scene features, our model predicts trajectories that capture the semantic intent while respecting spatial boundaries.

As shown in the Fig.~\ref{fig:qual_sup_pred_3_4} (bottom), the predicted motion paths accurately curve around obstacles and navigate through doorways, maintaining physically plausible distances from surrounding structures.
In contrast, our ablation variant, which lacks explicit 3D scene conditioning, generates motions that are kinematically coherent but environmentally inconsistent.
EgoLM often produces trajectories that drift into obstacles, failing to align the agent's orientation with the available free space.
Similarly, while UniEgoMotion generates semantically plausible motions, it ignores the structural layout, leading to collisions.
These results collectively highlight the superiority of Ego3DLM in forecasting future behaviors that are not only semantically aligned with the past egocentric video but also geometrically consistent with the complex 3D environment.

\end{document}